\documentclass[runningheads]{llncs}
\usepackage{graphicx}

\usepackage{tikz}
\usepackage{comment}
\usepackage{amsmath,amssymb} 
\usepackage{color}

\usepackage[accsupp]{axessibility}  

\usepackage[width=122mm,left=12mm,paperwidth=146mm,height=193mm,top=12mm,paperheight=217mm]{geometry}
\usepackage{booktabs}

\begin{document}
\pagestyle{headings}
\mainmatter
\def\ECCVSubNumber{7144}  

\title{HDR Reconstruction from \\ Bracketed Exposures and Events} 

\titlerunning{HDR from Bracketed Exposures and Events}
%
\author{Richard Shaw\and
Sibi Catley-Chandar\and
Ale\v{s} Leonardis \and \\
Eduardo P\'{e}rez-Pellitero}
\authorrunning{R. Shaw et al.}
%
\institute{Huawei Noah's Ark Lab, London, UK\\
\email{\{richard.shaw, sibi.catley.chandar, ales.leonardis, e.perez.pellitero\}@huawei.com}}
\maketitle

\begin{abstract}
Reconstruction of high-quality HDR images is at the core of modern computational photography. Significant progress has been made with multi-frame HDR reconstruction methods, producing high-resolution, rich and accurate color reconstructions with high-frequency details. However, they are still prone to fail in dynamic or largely over-exposed scenes, where frame misalignment often results in visible ghosting artifacts. Recent approaches attempt to alleviate this by utilizing an event-based camera (EBC), which measures only binary changes of illuminations. Despite their desirable high temporal resolution and dynamic range characteristics, such approaches have not outperformed traditional multi-frame reconstruction methods, mainly due to the lack of color information and low-resolution sensors. In this paper, we propose to leverage both bracketed LDR images and simultaneously captured events to obtain the best of both worlds: high-quality RGB information from bracketed LDRs and complementary high frequency and dynamic range information from events. We present a multi-modal end-to-end learning-based HDR imaging system that fuses bracketed images and event modalities in the feature domain using attention and multi-scale spatial alignment modules. We propose a novel event-to-image feature distillation module that learns to translate event features into the image-feature space with self-supervision. Our framework exploits the higher temporal resolution of events by sub-sampling the input event streams using a sliding window, enriching our combined feature representation. Our proposed approach surpasses SoTA multi-frame HDR reconstruction methods using synthetic and real events, with a 2dB and 1dB improvement in PSNR-L and PSNR-$\mu$ on the HdM HDR dataset, respectively.
\end{abstract}

\section{Introduction}

\begin{figure*}[tbh!]
  \centering
  \includegraphics[width=\linewidth]{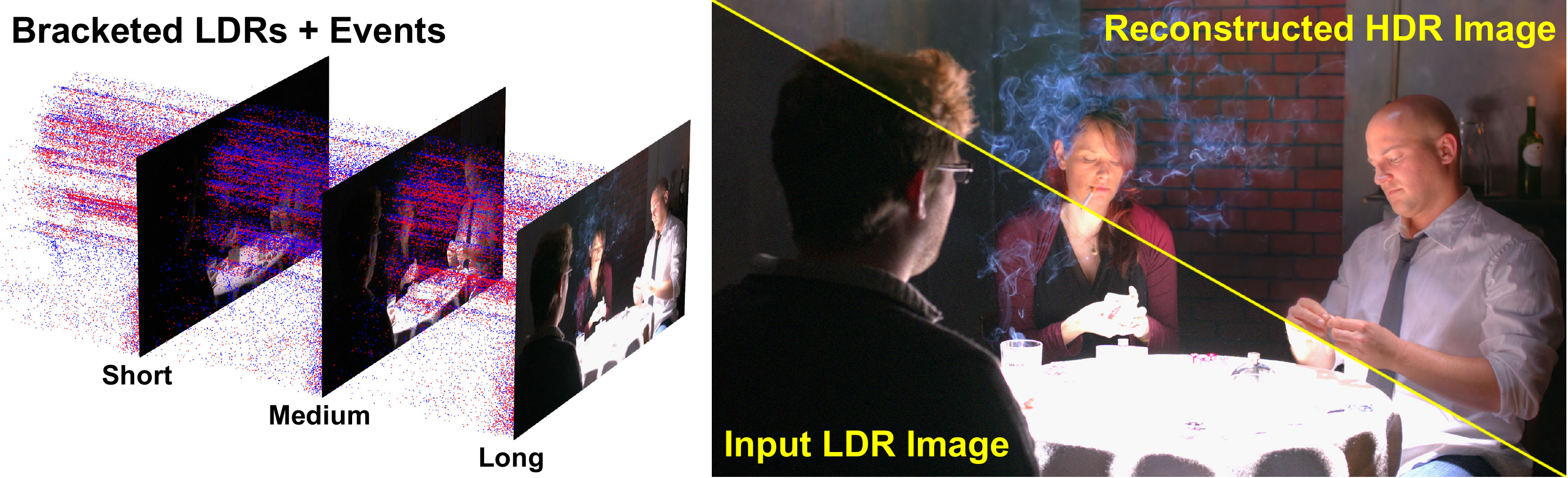}
  \caption{Our learning-based approach produces high-quality HDR images, leveraging both bracketed exposure LDR images from a conventional camera and event streams from an event-based camera. The two modalities provide complementary visual information: events provide higher frequency and dynamic information while RGB LDRs provide color and fine details. Left: the inputs to our model are a sequence of low-frequency LDRs captured with different exposures (short, medium and long) and an event stream, where the color denotes the event polarity (red is positive and blue negative). Right: Input LDR reference frame and our HDR reconstruction result.}
  \label{fig:teaser}
\end{figure*}

High dynamic range (HDR) imaging techniques extend the luminance range capturable beyond conventional or low dynamic range (LDR) cameras. Numerous HDR imaging strategies have been developed over recent years by the vision and graphics communities, as summarized by \cite{Sen2016,Wang2021DeepLF}. Traditionally, HDR methods involve capturing multiple LDR images with varying exposure values, termed bracketed exposures, and then merging them with different weights to reconstruct an HDR image \cite{Debevec2008}. The seminal work of \cite{Kalantari2017} extends multi-frame fusion to dynamic scenes by curating paired dynamic input LDR images with static ground truth HDR labels, and training end-to-end deep neural networks. This paradigm has been very successful over recent years, and as of now defines the state-of-the-art in HDR reconstruction~\cite{Liu2021,Niu2021}. However, these methods have limitations intrinsic to the camera sensor and bracketing strategy: frames generally need to be aligned to a reference frame due to the sequential capturing. Finding frame correspondences is challenging and might be affected by e.g.\,non-rigid motion or disoclussions, which results in motion-related artifacts. Furthermore, the dynamic range available per frame is limited, and thus each frame within the exposure bracket will inevitably miss some parts of the scene (i.e.\,under- or over-exposed). This is especially problematic for the reference frame, as alignment to areas that suffer information loss is not well defined in terms of photometric loss. Other approaches reconstruct from only a single LDR image; an ill-posed problem where texture details in poorly-exposed regions are hallucinated from neighboring areas or priors learned through deep neural networks \cite{Eilertsen2017}. Multi-frame methods, however, continue to out-perform single image approaches \cite{PerezPellitero2021}.

Recently, event-based cameras (EBC), or neuromorphic cameras, have garnered significant attention from researchers due to their unique properties distinct from conventional frame-based cameras. EBCs are novel bio-inspired sensors presenting a paradigm shift in how visual information is acquired; while a standard frame-based camera captures intensity images at a fixed frame rate, event cameras detect changes in per-pixel log intensity $L = \log (I)$ (brightness) asynchronously. An event $E_i = \left(x_i, y_i, t_i, p_i \right)$ is triggered at pixel $(x_i,y_i)$ at time $t_i$ when the brightness increment since the last event at that pixel, i.e. $\Delta L(x_i, y_i, t_i) = L(x_i, y_i, t_i) - L(x_i, y_i, t_i-\Delta t_i)$ exceeds a contrast threshold $\pm C$, i.e. $\Delta L(x_i, y_i, t_i) = p_i C$, where $C>0$ and polarity $p_i \in \{+1, -1\}$ is the sign of the brightness change \cite{Gallego2022}. This framework enables capture at very high temporal resolution (in the order of $\mu$s), with high dynamic range (140dB vs 60dB), and low power consumption, making them appealing for HDR applications \cite{Gallego2022}. Despite these advantages, EBCs generally have a low spatial resolution and typically only record grayscale information and thus have so far struggled to produce high-resolution, colour-accurate and artifact-free image reconstructions.

In this work, we address these limitations and exploit the strengths of each capturing modality. As Fig.~\ref{fig:teaser} shows, we propose a multi-modal HDR imaging method combining high-quality RGB bracketed exposures from a frame-based camera and high temporal resolution and dynamic range events from an event-based camera. The main contributions of this paper are: \textbf{(1)}~A multi-modal HDR architecture that for the first time combines bracketed exposure LDRs and event streams from an EBC trained in an end-to-end manner.
\textbf{(2)}~An event-to-image distillation module that transforms event features into the image feature space without needing an intermediate intensity image, trained with self-supervision from features extracted from corresponding LDRs.
\textbf{(3)}~An event sampling mechanism using a sliding window that leverages their high temporal resolution by extracting subsets of events and spatially aligning them in feature space.

\section{Related Work}

\noindent\textbf{HDR Reconstruction from Bracketed LDRs} Bracketed HDR methods capture a series of differently exposed LDRs and merge them in a weighted fashion \cite{Debevec2008}. In dynamic scenes, the LDR images must be aligned to the reference frame, commonly achieved using optical flow, and then processed using an HDR reconstruction network. \cite{Kalantari2017} pioneered this two-stage process, using classical optical flow \cite{Liu2009} to align low- and high-exposure images to the medium frame and train a CNN to merge and correct alignment errors, supervised by the ground truth HDR image. \cite{Wu2018} used a similar approach, but instead of optical flow they compute a homography for background alignment, relying on the CNN to correct foreground motion implicitly. \cite{Yan2019} also followed this pipeline but introduced attention to suppress undesired information from the LDRs before merging, e.g. misalignments and badly-exposed regions. \cite{Peng2018} and \cite{Prabhakar2019} replaced classical optical flow with learning-based approaches, e.g. FlowNet \cite{Dosovitskiy2015}, and \cite{Prabhakar2020} save computation by computing flow at a low resolution and upscaling. Despite substantial performance improvements, these methods still suffer ghosting, particularly for fast-moving objects and saturated regions. Recent developments include \cite{Niu2021} proposing a GAN-based approach, and \cite{Prabhakar2021} using a weakly supervised training strategy. Most relevant to ours, \cite{Liu2021} introduced deformable convolutions for alignment, which we adopt in our work, and won the 2021 NTIRE HDR challenge \cite{PerezPellitero2021}.
\\

\noindent\textbf{Intensity Reconstruction from Events} Numerous works have explored intensity image reconstruction solely from event data; since events can provide additional scene details in regions that may be badly exposed in LDRs. \cite{Belbachir2014} was one of the first to explore this; however, it was restricted to known camera motion, e.g. a rotating event camera for panoramic imaging. \cite{Bardow2016} advanced to generic camera motion, estimating joint intensity images and optical flow through cost function minimization. In the seminal work E2Vid, \cite{Rebecq2019} was among the first to employ a learning-based approach, utilizing a recurrent neural network (RNN) for video reconstruction. Other approaches include \cite{Zou2021} using an RNN, and \cite{Isfahani2019} employing a conditional generative adversarial network. Extensions to \cite{Rebecq2019} include: reducing network parameters and complexity \cite{Scheerlinck2020}, improving simulated event data generalization \cite{Stoffregen2020}, and enhancing HDR video temporal consistency by incorporating optical flow \cite{Zhu2018}. 

Although intensity reconstruction from events has progressed considerably, results are typically low resolution, grayscale, and exhibit artifacts with limited motion. Moreover, events primarily reflect edge information, and learning-based approaches must hallucinate details in textureless regions. \cite{MostafaviIsfahani2020,Wang2020,Wang2021} proposed reconstructing high-resolution HDR images from low-resolution events, and \cite{Wang2021DualTL} tried to learn more robust event representations by jointly learning HDR images with downstream tasks such as semantic segmentation and depth estimation via knowledge distillation \cite{Wang2021KnowledgeDA}. However, it remains that these methods suffer in dealing with event sparsity and fail to produce highly-detailed color reconstructions compared to image-based HDR methods.
\\

\noindent\textbf{Event-guided HDR Reconstruction} Rather than reconstructing intensity images solely from event data, previous approaches have used events to guide the LDR to HDR mapping in different ways. Notably, \cite{Han2020} proposed a multi-modal camera system and learning framework for HDR from a single LDR image and an intensity map generated from events. However, their method has two main drawbacks. First, the intensity map is generated from events using the off-the-shelf reconstruction network E2Vid \cite{Rebecq2021}. Therefore, they do not optimize HDR reconstruction from the LDR and events end-to-end, resulting in a point of model failure. Second, a single LDR limits its ability to handle scenes with extreme brightness ranges as single-LDR methods perform worse than multi-frame approaches. These two aspects hinder the algorithm's performance, which suffers from noticeable color artifacts in over-exposed regions, and its quantitative performance falls short when compared to SoTA multi-frame HDR methods that use a conventional camera only. Our method solves these issues using an end-to-end HDR network that learns from LDRs and events jointly. Specifically, we enhance bracketed multi-frame HDR reconstruction with multiple event streams, enabling the processing of more complex scenes. Moreover, we leverage information between the events and LDRs using knowledge distillation. Our unified HDR framework directly fuses images and events in the feature domain without relying on the intermediate step of event intensity image generation.

\section{Proposed Method}

Given a sequence of $n$ LDR images with different exposure values $\{I_1, I_2, ..., I_n\}$ captured at timestamps $\{t_1, t_2, ..., t_n\}$ and a stream of input events $\{E_i\}_{t_0}^{t_n}$ our aim is to reconstruct a single HDR image $H$ which is aligned to the reference frame $I_{ref}$ at timestamp $t_{ref}$. In our implementation, we use three input LDR images corresponding to short, medium and long exposures, specifying the middle frame as the reference $I_{ref} = I_2$. To generate inputs to our model, we follow \cite{Kalantari2017,Wu2018,Yan2019} forming a linearized image $L_i$ for each $I_i$ as follows: $L_i = I_{i}^{\gamma} \mathbin{/} T_i, i=1,2,3$,
where $T_i$ is the exposure time of image $I_i$. Setting $\gamma = 2.2$ approximates the inverse of gamma correction, while dividing by the exposure time adjusts all the images to have consistent brightness. 

Events and LDRs are acquired simultaneously but at different frequencies, i.e. LDRs are acquired at a lower frequency $\{ t_{1}, t_2, t_{3} \}$ and events at a higher frequency $E_i \in [t_{0}, t_{3}]$ where $t_0$ is the beginning of the captured event stream before the first LDR is acquired. The events provide additional information in-between the low-frequency LDRs and different parts of the event stream correspond to the acquisition of a different LDR image, therefore we partition the input event stream into three chunks corresponding to the LDR timestamps: $\{E_1, E_2, E_3\} = \{ E_{t_0 \rightarrow t_{1}}, E_{t_{1} \rightarrow t_2}, E_{t_2 \rightarrow t_{3}} \}$. Therefore, our proposed network $g$ can be defined as
$\hat{H} = g
\left(
\{I_i\}, \{L_i\}, \{E_i\}
; \theta
\right)$,
where $\hat{H}$ denotes the reconstructed HDR image and $\theta$ the network parameters.

Following \cite{Liu2021}, instead of concatenating the inputs and processing jointly, we use a multi-branch input pipeline where each input modality is processed separately before fusion. Specifically, for the LDR images $\{I_i\}$ we learn attention feature maps with a spatial attention module $\mathcal{A}$ to suppress misaligned and badly-exposed regions in the LDRs. The gamma-corrected linear images $\{L_i\}$, are processed using a pyramidal, cascading and deformable (PCD) alignment module $\mathcal{P}^L$ to handle scene or camera motion. We extend the method for events $\{E_i\}$, using a separate PCD module $\mathcal{P}^{E}$ to spatially align event features to the reference timestamp. Finally, a feature distillation module $\mathcal{D}$ is used to transform intermediately sampled events $\{E_{j}\}$ into the image feature space. Therefore, more accurately our end-to-end network $g$ can be described as:
\begin{equation}
    \hat{H} = g
    \left(
    \mathcal{A}(I_i), 
    \mathcal{P}^L(L_i), 
    \mathcal{P}^E(E_i),
    \mathcal{D}(E_j)
    ; \theta
    \right)
\end{equation}
\subsection{Network Architecture}

\begin{figure*}[tbh!]
  \centering
  \includegraphics[trim={0cm 4.5cm 0cm 0cm},clip,width=\linewidth]{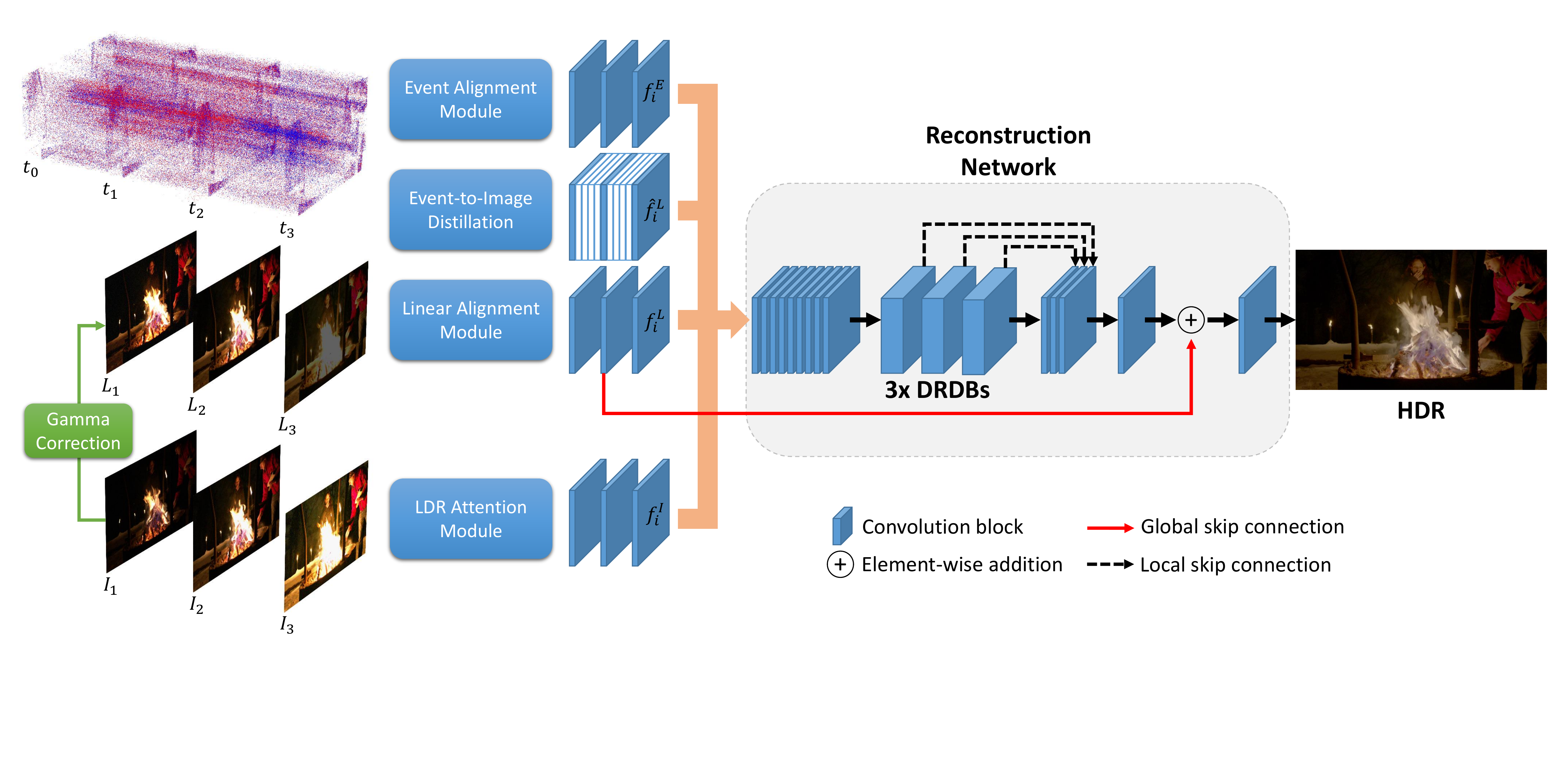}
  \caption{Model architecture. It accepts bracketed LDR images and event streams as input, extracting features from each modality with convolutional layers. LDRs are fed through an attention module, gamma-corrected linear images and events are spatially aligned using separate pyramidal deformable convolution (PCD) alignment modules. Events are temporally sub-sampled and translated to intermediate pseudo-image features using a feature distillation sub-network. Finally, the merging network, comprising residual dense blocks, fuses the aforementioned input branches for HDR reconstruction.}
  \label{fig:network}
\end{figure*}

In this section, we provide an overview of our approach (shown in Fig. \ref{fig:network}) by explaining different stages of our system. Our architecture is composed of five main components: 1) a spatial attention module $\mathcal{A}$ for LDR images, 2) an alignment module $\mathcal{P}^L$ for gamma-corrected linear images, 3) an alignment module for input event features $\mathcal{P}^E$, 4) an event-to-image feature distillation and alignment network $\mathcal{D}$, and 5) a fusion and HDR reconstruction network.

\subsubsection{1) LDR Spatial Attention Module}
Following \cite{Yan2019}, a spatial attention module $\mathcal{A}$ learns attention maps from the three input LDR images. Given LDRs $\{I_1, I_2, I_3\}$ with shape $H \times W \times 3$, we extract their LDR features using a single convolutional layer. For each non-reference LDR image ($I_i \neq I_2$), we concatenate the features with the reference image features $f^{I}_{ref}$ as input to the attention module, comprising two convolutional layers and a sigmoid function, generating an attention map in the range 0-1. Element-wise multiplication of LDR features with the corresponding attention map generates spatially attenuated features for each LDR image: $f^{I}_i = \mathcal{A} \left(I_i, I_{ref}\right), i=1,3$.

\subsubsection{2) Linear Image Alignment Module}
Following \cite{Liu2021}, we use a PCD alignment module $\mathcal{P}^L$ to align gamma-corrected linear images $\{L_i\}$ at the feature level. As shown by \cite{Chan2021UnderstandingDA}, alignment at the feature level is typically better than at the image level, and we accomplish this with deformable convolution \cite{Dai2017DeformableCN}. As in \cite{Wang2019EDVRVR}, we extract multi-scale pyramids of features using strided convolutions for each $\{L_i\}$ and then perform deformable alignment to the reference features $f^L_{ref}$ at each scale: $f^{L}_i = \mathcal{P}^L \left( L_i, L_{ref} \right), i=1,3$.

\subsubsection{3) Event Alignment Module}
\label{sec:event_alignment}
For the event modality we employ a separate PCD alignment module to perform spatial alignment in the event feature domain. \cite{Zou2021} demonstrated that alignment of events using deformable convolutions is effective for the task of intensity image reconstruction. In detail, we align the partitioned input event streams $E_1 = \{E_{t_0 \rightarrow t_1}\}$ and $E_3 = \{E_{t_2 \rightarrow t_3} \}$ to the events accumulated during the capture of the reference frame corresponding to the reference timestamp $E_{ref} = \{E_{t_1 \rightarrow t_2}\}$: $f^{E}_i = \mathcal{P}^E \left( E_i, E_{ref} \right), i=1,3$.

\subsubsection{4) Event-to-Image Distillation Module}
\label{sec:event-to-image}
To further leverage complementary information between events and images, we introduce a novel feature distillation module $\mathcal{D}$ that learns to transform event features into the image feature domain, since our end goal is to predict an HDR image. Due to higher rate at which events are captured, the events provide extra information in-between the low-frequency LDR images. Events accumulated during different parts of the event stream correspond to the acquisition of a different LDR image, i.e.\,three partitions of the event stream $\{E_1, E_2, E_3\} = \{ E_{t_0 \rightarrow t_1}, E_{t_{1} \rightarrow t_2}, E_{t_2 \rightarrow t_{3}} \}$ correspond to the gamma-corrected linear images $\{L_1, L_2, L_3\}$ at timestamps $\{t_1, t_2, t_3\}$. Therefore, for the network to learn to translate event features into linear image features, we can apply a self-supervising $\ell_2$ loss between the extracted features from the events $f^{E}_{i}$ and the corresponding linear image features $f^{L}_{i}$ at each timestamp:
\begin{equation}
  \mathcal{L_D} = 
  \sum^3_{i=1}
  \left(
  f^{E}_{i} - 
  \mathrm{sg} \left( f^{L}_{i} \right)
  \right)^2
  \label{eq:featureloss1}
\end{equation}
where sg(·) indicates a stop-gradient that treats its argument as an constant with zero derivative, preventing it from influencing the loss gradient during backpropagation, i.e. the learnt linear image features $f^{L}_{i}$ are treated as a self-supervising label. This encourages the domain transfer of event features into image features. Since we extract multi-scale pyramids of event and image features, we apply this loss at each scale $s \in S$ of the feature pyramid:
\begin{equation}
  \mathcal{L_D} =
  \sum^S_{s=1}
  \sum^3_{i=1}
  \left(
  f^{E}_{i,s} - 
  \mathrm{sg} \left( f^{L}_{i,s} \right)
  \right)^2
  \label{eq:featureloss2}
\end{equation}

\begin{figure}[ht]
  \centering
  \includegraphics[trim={0cm 1cm 0cm 1cm},clip,width=\linewidth]{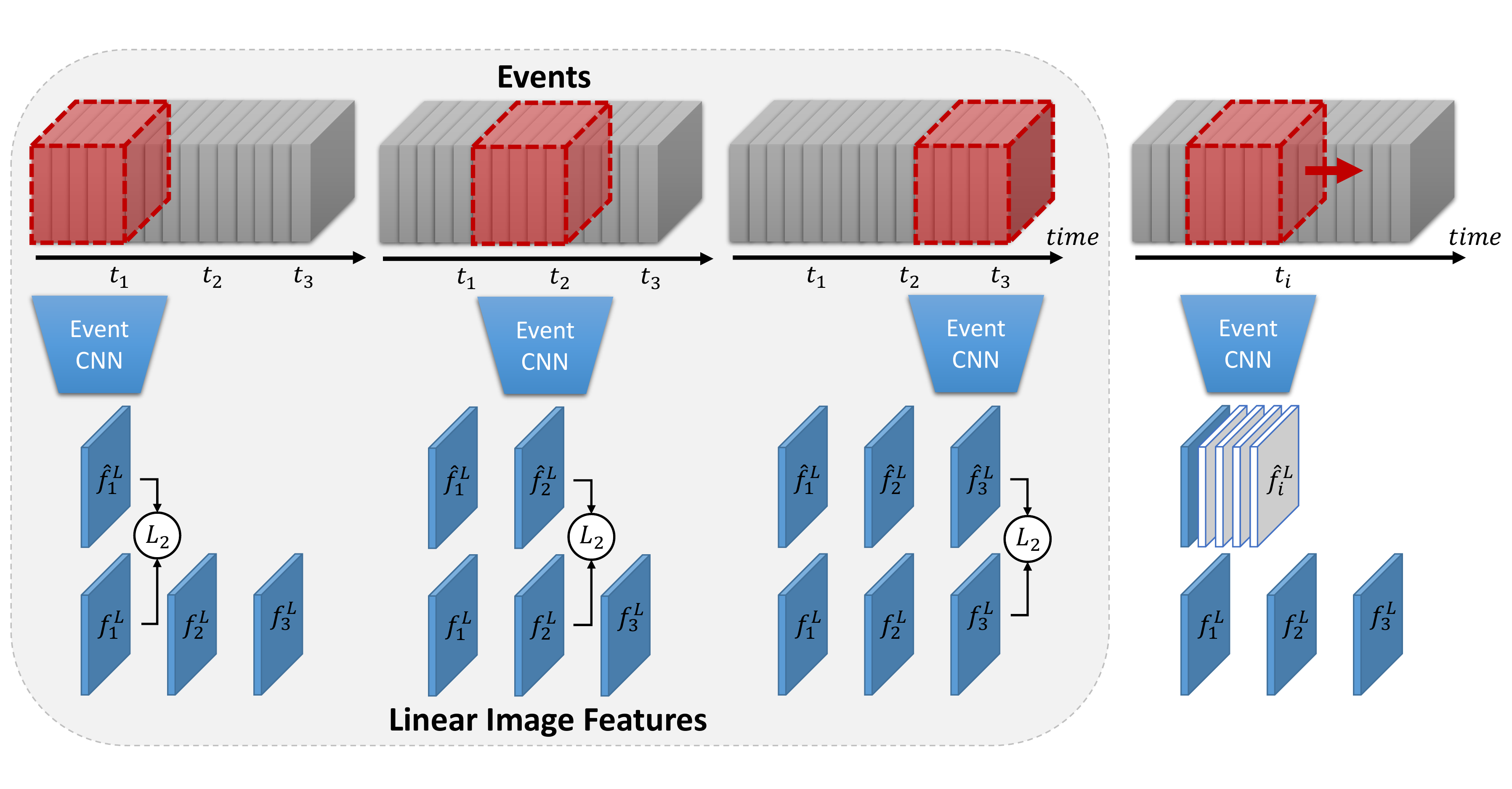}
  \caption{Our event-to-image feature distillation module transforms event features into image domain features. Left: our self-supervision training strategy employs a loss on the corresponding LDR features. Right: intermediate features are generated by sub-sampling events with a sliding window.}
  \label{fig:eventsampler}
\end{figure}

By learning an event-to-image distillation sub-network, self supervised by linear image features $f^{L}_{i}$, we can exploit the higher temporal resolution of the event data by sub-sampling the input event stream using a sliding window approach. Sampling chunks of events in-between the LDR keyframes and passing these through the learnt distillation network enables us to predict intermediate pseudo-image features $\hat{f}^{L}_{i}$ (event features transformed to image domain), thereby enriching our combined image and event feature representation. This  process is shown schematically in Fig. \ref{fig:eventsampler}.

\subsubsection{5) HDR Reconstruction Sub-network}
The feature maps from the four aforementioned input branches (LDRs after attention, aligned linear images, aligned events, and event-to-image features) are concatenated together as the input to the HDR fusion sub-network. The fusion sub-network follows good practices common in recent literature, e.g.~\cite{Liu2021,Yan2019}, consisting of three dilated residual dense blocks (DRDBs), with dilated convolutions \cite{Yu2017DilatedRN} to increase the receptive field, and global and local skip connections. As shown by the ablation results in section \ref{sec:results}, we find the addition of each of these submodules leads to improved HDR quality with less ghosting and better detail reconstruction.

\subsubsection{Loss Function}
As HDR images are not viewed in the linear domain, we follow previous work~\cite{Liu2021,Yan2019} and use the $\mu$-law to map from the linear
HDR image to the tonemapped image:
\begin{equation}
  \mathcal{T}(H) = 
  \frac{\log(1 + \mu H)}
  {\log(1 + \mu)}
  \label{eq:tonemap}
\end{equation}
where $H$ is the linear HDR image, $\mathcal{T}(H)$ is the tonemapped image and $\mu = 5000$. We then estimate the $\ell_1$-norm between the prediction and the ground truth HDR image as follows: $\mathcal{L}_{HDR} = {|| \mathcal{T}(\hat{H}) - \mathcal{T}(H) ||}_1$.
In addition to the tonemapped HDR reconstruction loss, as discussed in section \ref{sec:event-to-image} we use a self-supervising loss given by Eq.~(\ref{eq:featureloss2}) on the features generated from our event-to-image feature distillation module. Therefore, the total loss is the sum of the HDR reconstruction loss and distillation losses: $\mathcal{L}_{total} = \mathcal{L}_{HDR} + \mathcal{L_D}$.

\subsection{Training Data}
\label{sec:training-data}
For training, we model bracketed exposure LDRs from ground truth HDR video frames and generate synthetic event data using ESIM: Event Camera Simulator \cite{Rebecq18corl}. We obtain ground truth HDR frames from the HdM HDR dataset \cite{Froehlich2014}, which contains sequences with varied scenes, lighting and motion. Following \cite{PerezPellitero2021}, using four scenes for validation/testing and 25 scenes for training, ensuring no scene overlap between training and testing/validation splits, we obtain 1500, 60 and 201 samples for training, validation and testing, respectively.
\\

\noindent \textbf{Image Formation Model}: \label{sec:imagemodel}
To generate bracketed LDR frames $\{I_i\}$ we use the pixel measurement model from \cite{Hasinoff2010}: $I_{i} = \min \left( \Phi T / g + I_0 + n, I_{max} \right)$,
where $\Phi$ is the scene brightness, $T$ exposure time, $g$ sensor gain, $I_0$ offset current, $n$ sensor noise and $I_{max}$ the saturation point. We approximate $\Phi$ by the ground truth HDR image and generate LDRs by modifying $T$ for any three consecutive frames.
\\

\noindent\textbf{Noise Model}: Following \cite{PerezPellitero2021}, we include a zero mean noise signal $n$ whose variance comes from three sources: photon noise, read noise, and ADC gain and quantization: $\mathrm{Var}(n) =  \Phi / g^2 + \sigma^2_{read} / g^2 + \sigma^2_{ADC}$.
\\

\begin{figure}[htbp!]
  \centering
      \includegraphics[trim={0cm 10.4cm 0cm 9cm},clip,width=1.0\linewidth]{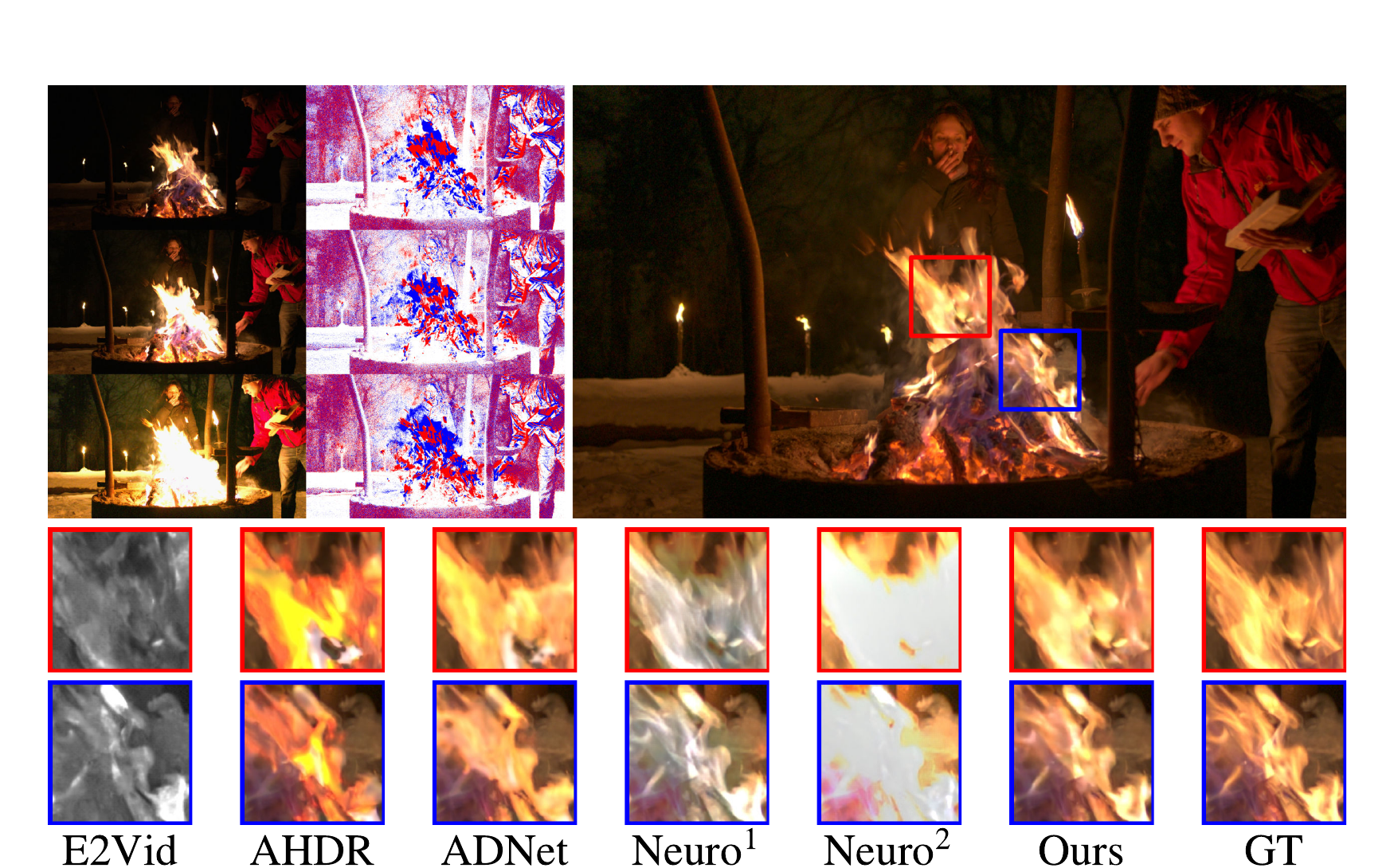}
    \includegraphics[trim={0cm 10.4cm 0cm 9cm},clip,width=1.0\linewidth]{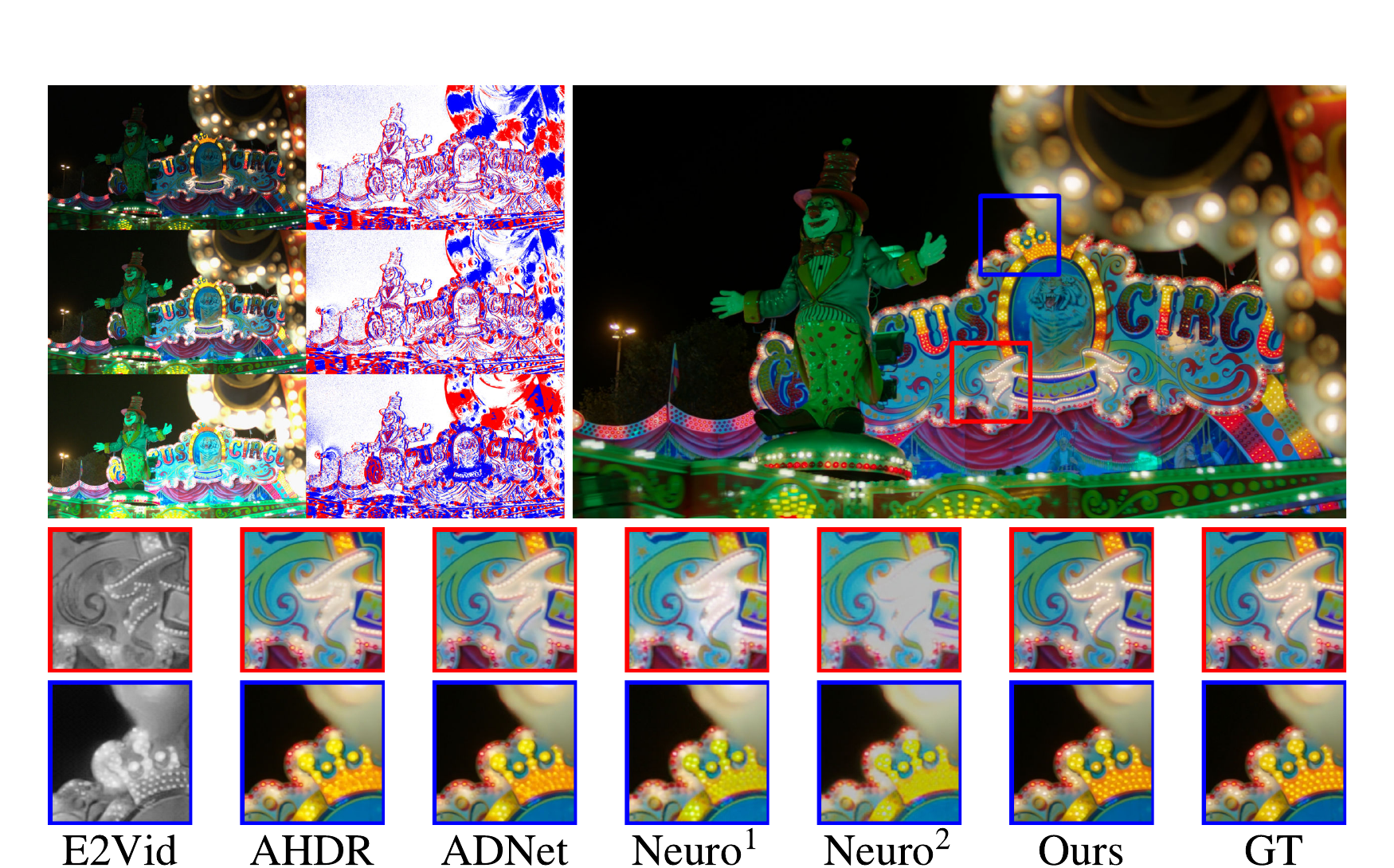}
  \caption{Qualitative results on the HdM test set with synthetically generated events. Top left: input LDRs and events, top right: the predicted HDR image using our method, and bottom row: comparison crops for each method.}
  \label{fig:synthetic1}
\end{figure}

\noindent\textbf{Event Generation Model}: To generate high-frequency events, we follow Vid2e \cite{Gehrig2020Vid2e} by temporally upsampling the HdM HDR video sequences using Super SloMo \cite{Jiang2018SuperSH}. In contrast to other works which use 8-bit LDR video as input, to generate realistic events that retain a high dynamic range, we first $\mu$-tonemap the ground truth HDR frames to the nonlinear domain using Eq.~(\ref{eq:tonemap}), as the pre-trained Super SloMo network was trained on tonemapped LDR videos. The resulting interpolated frames are converted back to the linear domain with the reverse tonemap transformation. ESIM then processes the upsampled frames with a contrast ratio of $C = 0.5$ generating binary event streams $\{E_i\} = \{(x_i, y_i, t_i, p_i)\}$. Examples of our synthetic event data are shown in Fig.~\ref{fig:synthetic1}.

\subsection{Implementation Details}

We implement our model in PyTorch. Each module consists of $3 \times 3$ convolutions, extracting 64 feature channels, and the network ends with a ReLU predicting an unbounded linear HDR image. During training, we randomly sample crops of size $256 \times 256 \times 3$ from the input LDR images and corresponding crops of $256 \times 256 \times 5$ from the input event voxel grids. Augmentations are applied consisting of random horizontal and vertical flipping and $90^\circ$ rotations. We train each model for 2000 epochs using a batch size of 16 and a learning rate of $10^{-4}$ with Adam optimizer. We employ a stepped learning rate schedule decaying by factor 10 every 500 epochs. At test time, the full resolution image is processed.
\\

\noindent\textbf{Event Representation}: To process events using a CNN, we discretize the time axis into $B$ bins. Following \cite{Zhu2018}, we extend the effective temporal resolution beyond $B$ by weighted accumulation of events. Given a set of $N$ events $\{E_i\}_{i=0,...,N-1} = \{(x_i, y_i, t_i, p_i) \}_{i=0,...,N-1}$, we scale the range of timestamps $\Delta t = t_{N-1} - t_0$ to the range $[0, B-1]$; each event distributing its polarity $p_i$ to the two closest spatio-temporal voxels: $E(x_i,y_i,t_i) = \sum_i p_i \max (0, 1-|t_i - t^*_i|)$,
where $t^*_i = \frac{B-1}{\Delta t} (t_i - t_0)$ is the normalized event timestamp. We use $B = 5$ as is typically used in the literature \cite{Rebecq2019}. Note, that increasing $B$ has limited influence on reconstruction quality at the expense of increased computational cost.

\section{Results}
\label{sec:results}

This section presents quantitative and qualitative results on synthetic and real event data, using PSNR metrics in the linear and tonemapped domains (PSNR-L and PSNR-$\mu$) and the HDR-VDP-2 (HV2) metric \cite{Mantiuk2011HDRVDP2AC}. We compare the results of our model to five other methods. Specifically, we compare two SoTA bracketed exposure image-based methods AHDR \cite{Yan2019} and ADNet \cite{Liu2021}. We also compare to an event-only method E2Vid \cite{Rebecq2019}. Finally, we compare to two Neuromorphic HDR \cite{Han2020} models (Neuro$^1$ and Neuro$^2$) which combine a single LDR image with an event intensity map. Neuro$^1$ follows the training procedure proposed in \cite{Han2020} which generates a synthetic event intensity map from the Poisson reconstruction of image gradients of the ground truth HDR image. Neuro$^2$ is the same model with the intensity map generated by E2Vid as proposed in their paper. We choose these models as they represent the best performing and most relevant models to the proposed method. Note that E2Vid is disadvantaged as it only uses event data and, therefore, cannot reconstruct color information reliably. Thus, for this model, we compute the metrics using grayscale ground truth HDR images.

\subsection{Synthetic Events Dataset}

Quantitative results using synthetic event data generated from the HdM HDR video test set are displayed in Table \ref{tab:tab_test} (left). Our method performs significantly better in all metrics over the other methods, with an over 2dB increase in PSNR-L and almost 1dB increase in PSNR-$\mu$ over the closest performing model ADNet. Note that E2Vid performs particularly poorly here because of its RNN architecture which is unstable for short sequences, and the fact that many frames in the HdM dataset have relatively little motion. The network falls into a failure case with a sparse event signal, unable to reconstruct the image properly. In contrast, our method can rely on the other input modality of the bracketed LDRs to maintain good HDR reconstruction performance even with a lack of events.



\setlength{\tabcolsep}{4pt}
\begin{table}
\begin{center}
\caption{Quantitative results on the HDM test set with synthetic events (left) and on DSEC test set with real events (right). Best performer on each dataset denoted in bold}
\label{tab:tab_test}
\begin{tabular}{lcccccr}
\toprule
& \multicolumn{3}{c}{HDM dataset} & \multicolumn{3}{c}{DSEC dataset}\\
\cmidrule(lr){2-4}\cmidrule(lr){5-7}
Method & PSNR-L & PSNR-$\mu$ & HV2  & PSNR-L & PSNR-$\mu$ & HV2 \\
\midrule
E2Vid \cite{Rebecq2019} & 22.44 & 14.68 & 38.90 & 12.04 & 10.84 & 64.29 \\
AHDR \cite{Yan2019} & 37.79 & 36.59 & 49.11 & 36.52 & 34.64 & 74.47\\
ADNet \cite{Liu2021} & 39.18 & 36.89 & 50.06  & 37.17 & 35.12 & 75.14\\
Neuro$^1$ \cite{Han2020} & 32.34 & 30.97 & 48.63 & 28.26 & 31.62 & 72.76 \\
Neuro$^2$ \cite{Han2020} & 27.45 & 25.98 & 36.72 & 22.58 & 24.76  & 68.92 \\
Ours & $\mathbf{41.81}$ & $\mathbf{37.84}$ & $\mathbf{55.79}$ & $\mathbf{38.13}$ & $\mathbf{35.83}$ & $\mathbf{76.65}$ \\
\bottomrule
\end{tabular}
\end{center}
\vspace{-3em} 
\end{table}
\setlength{\tabcolsep}{1.4pt}

\subsection{Real Events Dataset}
As there are no publicly available real-world datasets containing both ground truth HDR image frames and events, to perform experiments on real event data, we make use of the DSEC dataset \cite{Gehrig2021DSECAS}. In DSEC, a car-mounted rig captures LDR video and hardware synchronized events. Due to field-of-view, optical center and resolution differences, the events are spatially aligned to the video frames using a calibration checkerboard. We choose well-exposed LDR video frames as pseudo ground truth images (with no clipped shadows or highlights). From these pseudo-HDR frames, we further degrade them by applying the exposure model, adding noise, clipping and quantizing to generate short, medium and long exposure LDRs as discussed in section~\ref{sec:training-data}. Corresponding events are extracted from the raw event stream corresponding to the timestamps of the selected frames.

Since our models were trained on synthetic events, perfectly aligned to the LDRs, there exists a domain gap between the synthetic and real event datasets. Therefore, we fine-tune each model equally on the real-world training sets for a small amount of time. Specifically, we train each model for a further 200 epochs. This is necessary for our multi-modal image and events model as in our synthetic training data the events are perfectly aligned to the image frames. However, in the real event dataset, because the events and images are captured from two different cameras with a small baseline (with different sized sensors, resolutions and focal lengths), alignment errors and interpolation artifacts exist in the data. Fine-tuning enables the networks to handle these slight misalignments errors. 

\begin{figure}[htbp!]
  \centering
      \includegraphics[trim={0cm 9.9cm 0cm 8cm},clip,width=0.97\linewidth]{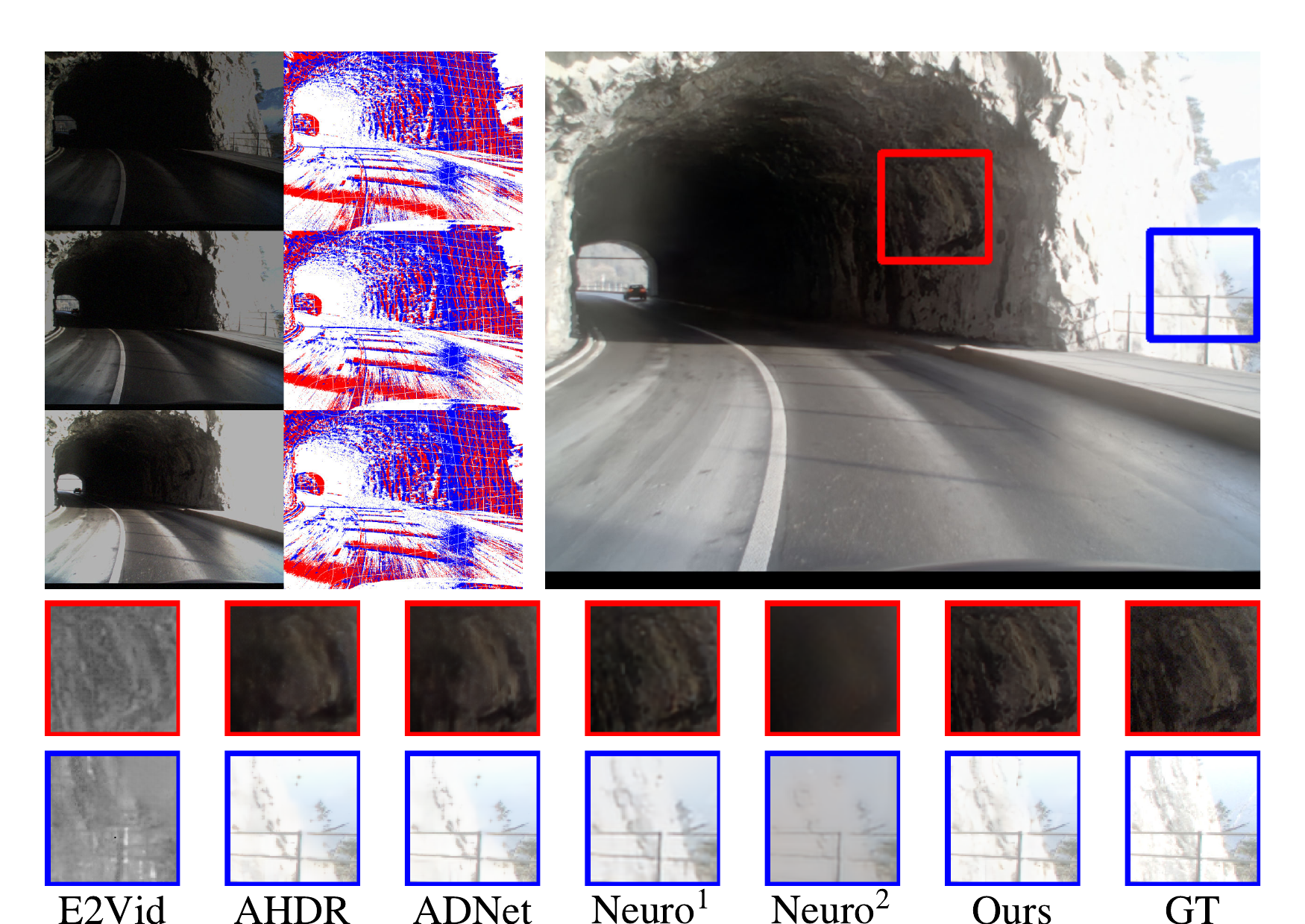}
    \includegraphics[trim={0cm 9.9cm 0cm 8cm},clip,width=0.97\linewidth]{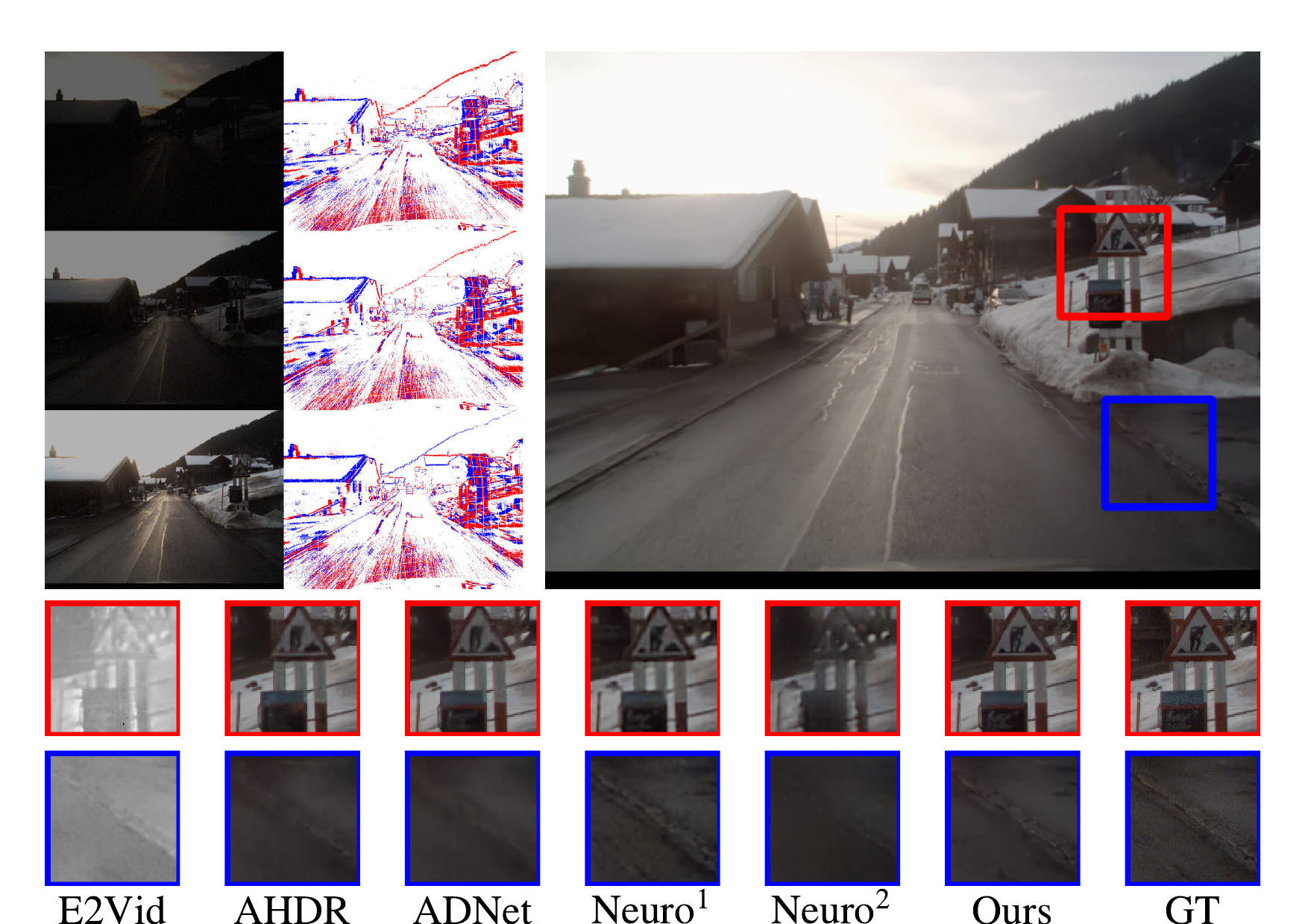}
  \caption{Qualitative results on the DSEC test set with real events. Top left: input LDRs and events, top right: the predicted HDR image using our method, and bottom row: comparison crops for each method.}
  \label{fig:real1}
\end{figure}

Quantitative results are shown in Table~\ref{tab:tab_test} (right). Similarly to the synthetic dataset, our approach outperforms all other methods in each metric, with large gaps of approximately $1$ dB and $0.7$ dB (PSNR-L and $\mu$) to the second best-performer ADNet.  Qualitative results are shown in Fig.~\ref{fig:real1}. We outline in the crops noticeably better texture reconstruction in fast moving objects (note that in this driving dataset, the most motion occurs towards the edges of the frame, with the middle region of the image being relatively static), such as the tunnel surface and the lane edge. High-contrast structures, e.g.~the traffic signal, are better recovered as well, and even in challenging regions of over-exposure object and scene structure is better preserved (e.g.~rock to sky transition).


\subsection{Ablation Studies}
\label{sec:ablation}

In this section, we perform several ablation studies examining the performance of each new component of our model. Quantitative results for each ablation on the HdM test set are shown in Table \ref{tab:ablation}. We train each model equally in the four following scenarios. First, \emph{Images-only} uses only bracketed LDRs as input and is equivalent to ADNet. Second, \emph{Images + Event alignment} is bracketed LDRs and the event alignment module discussed in section \ref{sec:event_alignment}. Third, \emph{Images + Event sub-sampling} is bracketed LDRs and sub-sampling of the event stream with alignment to the reference without feature distillation. And fourth, \emph{Images + Event-to-image distillation} is our complete model with sub-sampled events passed through the distillation module discussed in section \ref{sec:event-to-image}. The ablation test results show that each model configuration provides an increase in both PSNR metrics, thus validating the effectivity of our model design. The best result comes from the \emph{Images + Event-to-image distillation} model, which demonstrates the advantage of using feature distillation from events to images, rather than simply aligning the sub-sampled events in the event-feature domain. 

\setlength{\tabcolsep}{4pt}
\begin{table}
\begin{center}
\caption{Ablation studies for different model components on the HdM test set. Refer to Sec.~\ref{sec:ablation} for more details.}
\label{tab:ablation}
\begin{tabular}{lccr}
\toprule
Method & PSNR-L & PSNR-$\mu$ & HV2\\
\midrule
Images-only & 39.18 & 36.89 & 50.06\\
+ Event alignment & 40.97 & 37.35 & 55.14\\
+ Event sub-sampling & 41.32 & 37.60 & 54.85\\
+ Event-to-image distill. & $\mathbf{41.81}$ & $\mathbf{37.84}$ & $\mathbf{55.79}$ \\
\bottomrule
\end{tabular}
\end{center}
\vspace{0 em} 
\end{table}
\setlength{\tabcolsep}{1.4pt}

\subsection{Discussion}
As demonstrated by the experimental results, our method that uses bracketed exposures and events displays a clear advantage over methods that only use only events, images, or a single LDR and events. Each method has its limitations. Event-only methods cannot reconstruct images well in mostly static scenes due to the sparsity of events. Bracketed image-only methods perform much better on static scenes but suffer in scenes with significant motion and over-exposed LDRs, making alignment difficult. The method of single LDR and events performs better than events alone, but as the event intensity map input does not contain RGB color information, it does not do well at recovering detail and colors in over-exposed image regions. Since this method relies on a single LDR, it suffers similarly to single-image HDR reconstruction methods. Our method performs the best overall because it uses the best of both worlds.

\section{Conclusions}

In this paper, we have presented a learning-based method to leverage event data from an event-based camera and bracketed LDR exposures from a conventional frame-based camera to improve HDR image reconstruction performance. To the best of our knowledge, this is the first work to consider combined HDR reconstruction from both events and bracketed exposure images, which has a number of benefits over event-only or image-only HDR reconstruction methods. Specifically, in the case of static scenes where event-only methods fail (no event signal), our method is able to resort back to the input LDRs and performs equivalently to state-of-the-art image-based methods such as ADNet. And in dynamic scenes, where image-based methods struggle to align details between LDRs, our method is able to leverage the high frequency event data and better align and reconstruct fine details. A particular strength of our method is handling dynamic highlights, e.g.~flashing lights, fast-moving textures, where image-based methods struggle to align them to the reference frame due to saturated over-exposed pixels in the LDRs. Spatially aligning events and images in the feature domain enriches our feature representation and leads to better reconstructions, and ultimately performing a event-to-image domain distillation allows the system to find an optimal feature space for both events and images without the need of a event-based intensity guide image. We have  validated our method on both synthetically and real-world data, and conducted ablation studies supporting our contributions. Our method obtains significant improvements over other SoTA algorithms in all the measured metrics and noticeably improved visual results.

%
%
\bibliographystyle{splncs04}
\bibliography{egbib}

\clearpage
\section{Supplemental Materials}

\subsection{Video Results}
We strongly encourage the reader to view the video results included in the supplemental materials. The video presents additional results for each model on the HDM dataset with synthetic events and the DSEC dataset with real events. In-place comparisons, rather than side-by-side, show the differences between each model's predictions and the ground truth better. We zoom in to each image to highlight significant differences and our model's noticeably better detail reconstruction. 

We also include a sample of video results compiled from consecutive frames from each dataset. Our model displays better video reconstruction results with less temporal flickering than the other methods. Note that we compare with model Neuro$^1$ (using a synthetic event intensity map as input, generated from the Poisson reconstruction of ground truth HDR image gradients) on the synthetic event dataset, while we compare with Neuro$^2$ (using E2Vid as input) on the real event dataset, as this provides a fairer comparison. We observe that Neuro$^2$ performs much worse because it is not trained event-to-end from raw events to HDR image output and the E2Vid input intensity map is often of poor quality due to event sparsity.


\subsection{Architecture Details}
Here we provide some additional details regarding the specific architecture we have implemented for our model.

\subsubsection{Inputs}
In our implementation, the input LDR images are scaled to the range 0-1, while the ground truth HDR images are left unbounded. The input event streams are converted to voxel grids, each with 5 temporal bins, which are then scaled such that the occupied (non-zero) voxels have zero mean and variance 1. 

\subsubsection{Feature Encoders}
\begin{figure}[ht]
  \centering
  \includegraphics[trim={0cm 0cm 0cm 0cm},clip,width=0.4\linewidth]{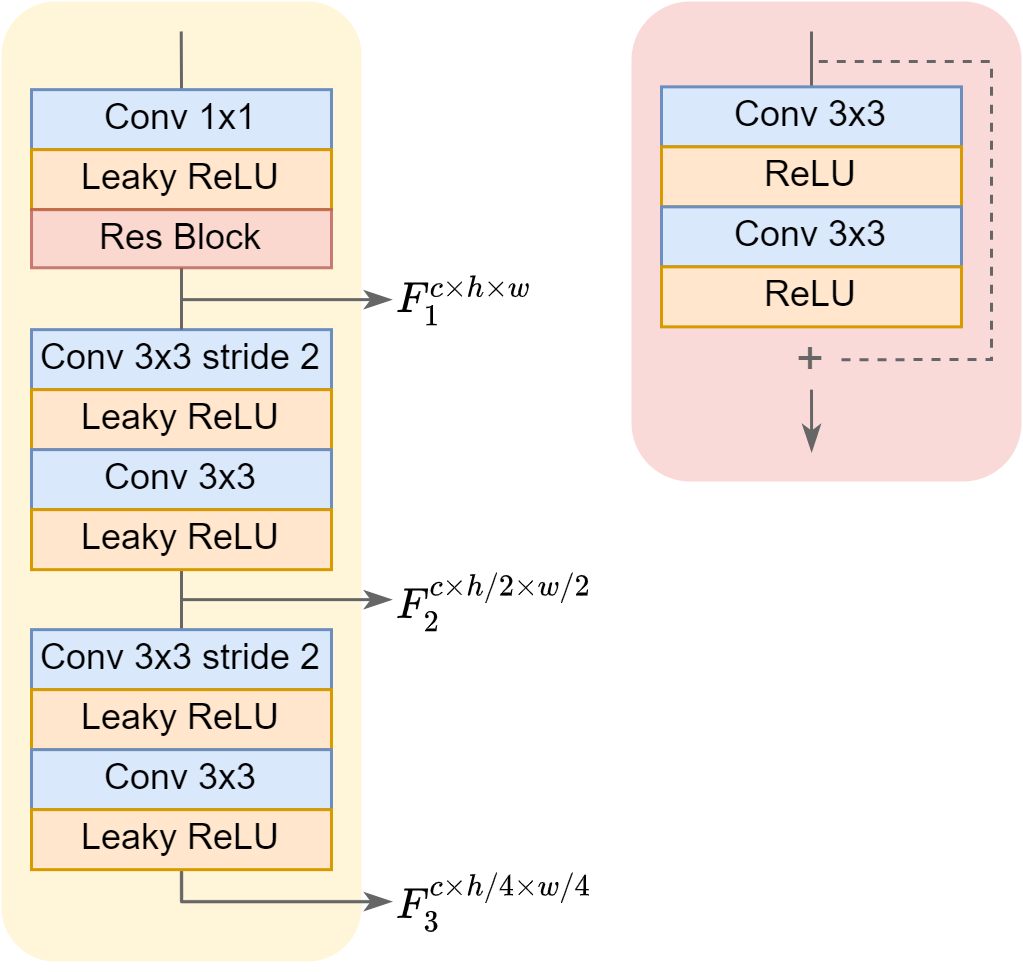}
  \caption{Feature encoder architecture.}
  \label{fig:encoder}
\end{figure}

The feature encoders for each processing module follow roughly the same architecture, as shown schematically in Fig. \ref{fig:encoder}. The only difference is the number of input channels depending on the input modality: 3 for images and 5 for events. Each encoder extracts a pyramid of features at 3 scales. First, initial features $F_1^{c \times h \times w}$ are extracted with a $1 \times 1$ convolution followed by a residual block to improve gradient flow. Then two convolutional blocks comprising a $3 \times 3$ convolution and strided convolution with stride 2 extract features $F_2^{c \times h/2 \times w/2}$ and $F_3^{c \times h/4 \times w/4}$ at half and quarter resolution respectively. Each convolutional filter extracts $c=64$ feature maps and is followed by either a ReLU or Leaky ReLU activation as shown in Fig. \ref{fig:encoder}. Each processing module has its own independent encoders -- no features are shared across the input modalities at this stage. In our experiments, we found performance to be better when each input is independently processed before fusing the features just before entering the HDR reconstruction backbone.

\subsubsection{Pyramidal Deformable Convolutional Alignment}
\begin{figure}[ht]
  \centering
  \includegraphics[trim={0cm 0cm 0cm 0cm},clip,width=0.9\linewidth]{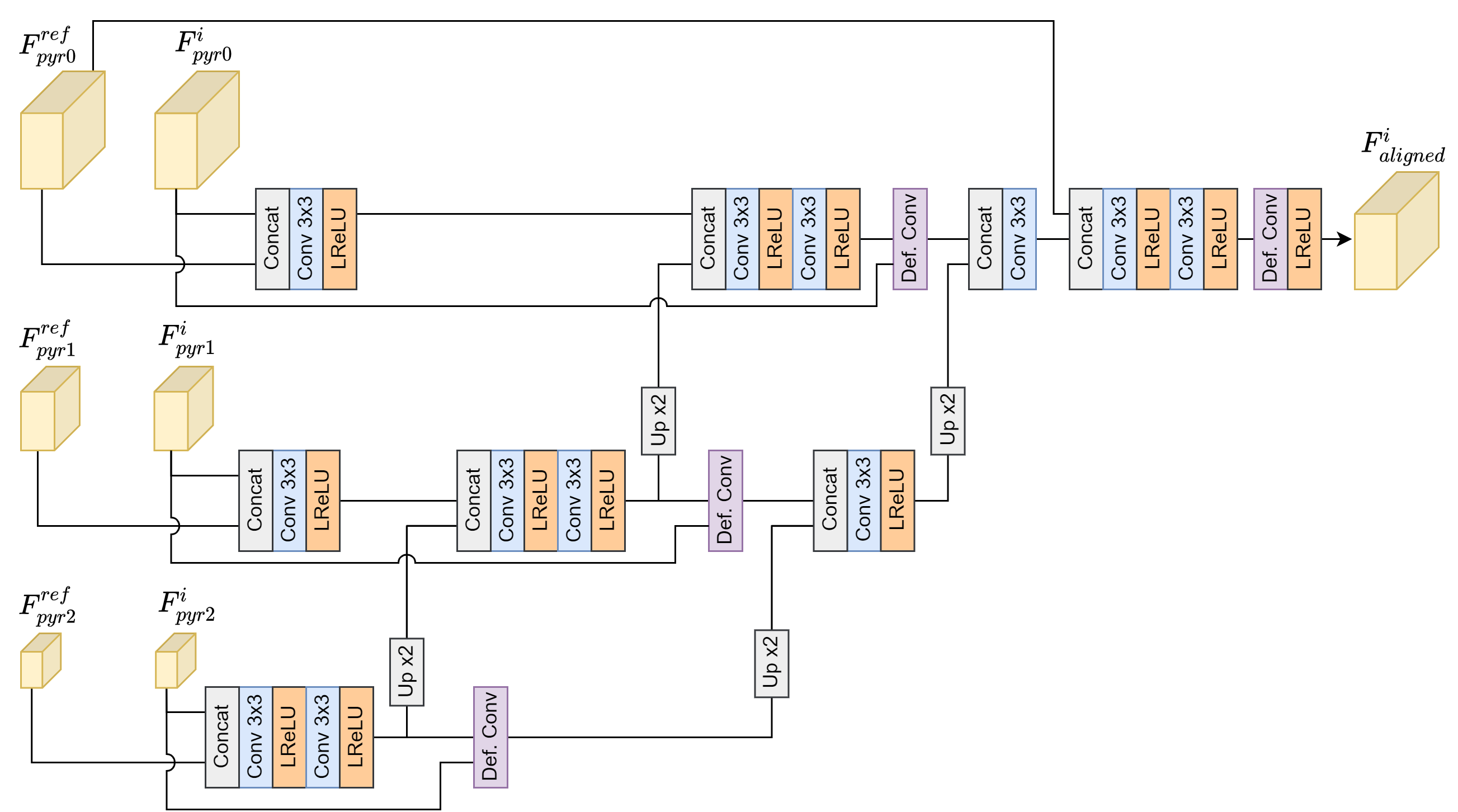}
  \caption{Feature alignment architecture.}
  \label{fig:spatial_alignment}
\end{figure}

As discussed in section \ref{sec:event_alignment}, deformable convolution modules are used to spatially align the event features and image features to the reference features. We use separate instances of the feature alignment module for processing each modality independently, i.e. features are not shared across the modalities. The architecture for the feature alignment module is shown in Fig. \ref{fig:spatial_alignment}, and again is roughly the same for both images and events. The $i^{th}$ input features $F^i$ are aligned to the reference features $F^{ref}$ at each scale of the feature pyramid $\{ pyr1, pyr2, pyr3 \}$. As shown in Fig. \ref{fig:spatial_alignment}, at each scale, the input and reference features are concatenated along the channel axis and passed through a series of $3 \times 3$ convolutions and Leaky ReLU activations. These features are then passed through deformable convolutions, essentially learning kernel offsets, and the output features from the lower scales are upsampled $\times 2$ using bilinear interpolation to guide the alignment of features at the higher resolutions.

Our experiments found that performance improved when the spatial alignment of events features and image features was performed separately, rather than concatenating the features of the different modalities prior to feature alignment. We also found that feature alignment using deformable convolutions outperformed the optical flow-based methods we tried. 

\subsubsection{HDR Reconstruction Backbone}
\begin{figure}[ht]
  \centering
  \includegraphics[trim={0cm 0cm 0cm 0cm},clip,width=0.85\linewidth]{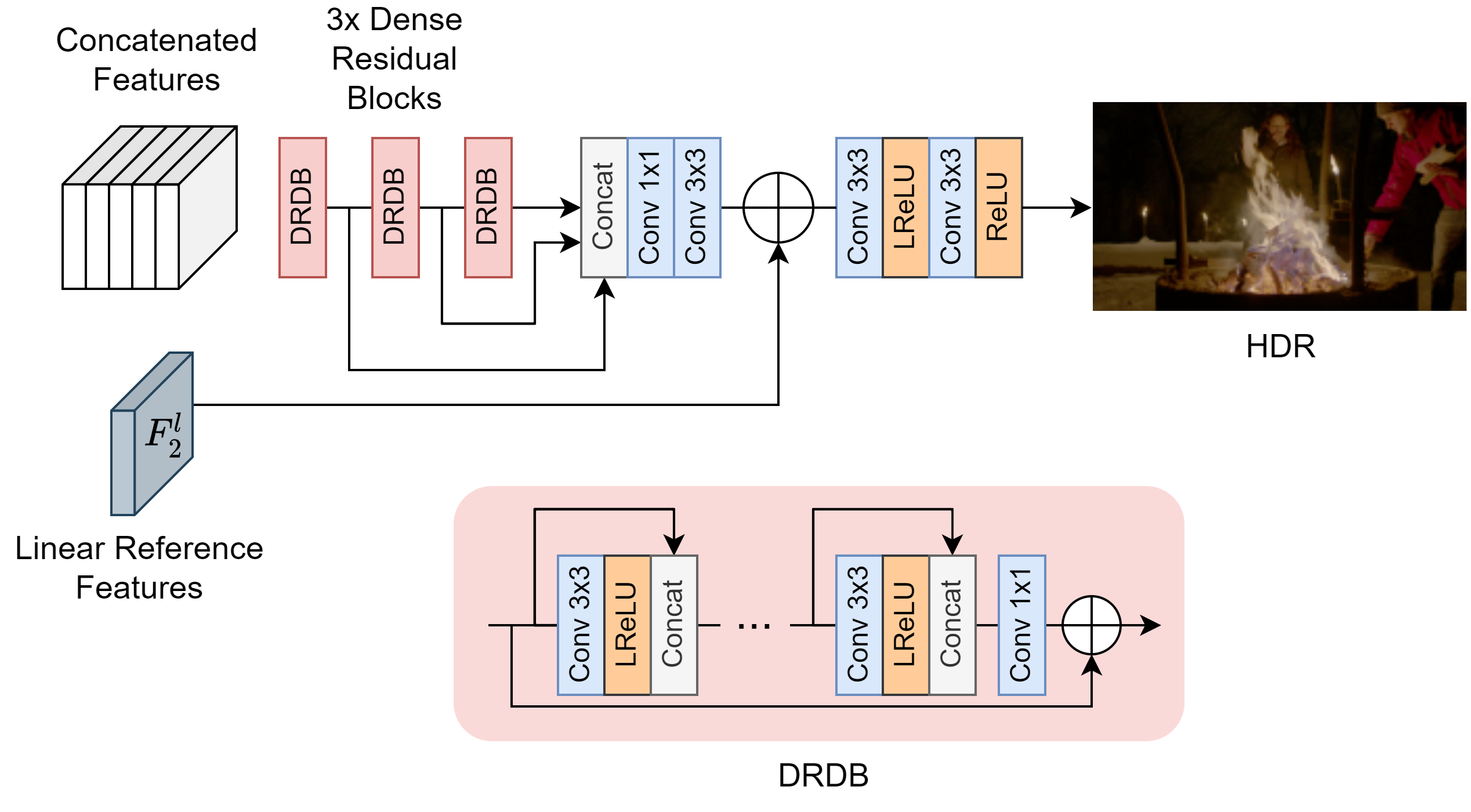}
  \caption{HDR reconstruction sub-network.}
  \label{fig:hdr_backbone}
\end{figure}

The architecture of the HDR reconstruction sub-network is shown in Fig. \ref{fig:hdr_backbone}. The network takes in the features from all previously processed submodules, containing both event and image features, which are then concatenated down the channel dimension. The input features are fused and reduced to 64 channels with a $3 \times 3$ convolution and then passed through a series of dilated residual dense blocks (DRDBs). Each DRDB comprises a $3 \times 3$ dilated convolution with dilation factor 2, followed by a $1 \times 1$ convolution and a local residual connection. 3 DRDBs are used with residual connections within each block, as shown in Fig. \ref{fig:hdr_backbone}. The linear reference features $F_2^L$ are added with a global skip connection. The reconstruction sub-network finishes with a series of convolutions and a final ReLU activation to predict an unbounded HDR image.

\subsection{Limitations}

\subsubsection{Model Complexity}
The main drawback of our method is the increased computational complexity of the model over image-only HDR methods. The increase in the number of model parameters stems mainly from the event-to-image feature distillation module, which sub-samples the event stream to generate intermediate feature inputs in-between the 3 LDR keyframes. The intermediate feature maps, each with a feature dimension of 64, are concatenated with the other input branches' features creating a large stack of combined features of size $N \times 64$, before being fused with a $3 \times 3$ convolutional filter down to just 64 features. Clearly, the size of the model depends on the number of sub-sampled features. In our implementation, we convert each input event stream to a voxel grid with 5 temporal bins and sample intermediate events using a sliding window along the time axis, resulting in 8 additional intermediate features maps in-between the 3 keyframes. In theory, due to the high temporal resolution of events, the event stream could be sampled many more times, but this would significantly increase the computational complexity of the model. We experimented with sampling fewer intermediate frames and also expanding the number of temporal bins to 10 and 20; however, we found that 5 bins provided the best performance with reasonable model size. Note that the optimal number of intermediately sampled features also depends on the capture frame rate (most scenes in the HDM dataset are captured at either 24fps or 25fps) and the amount of movement within the scene; scenes with significant or complex motion between the keyframes would require many more sampled features than a primarily static scene. Choosing the appropriate number of samples for the particular scene warrants further investigation. Furthermore, our model size could be reduced using any number of model size reduction techniques, for instance, model distillation by training a smaller network with supervision from this larger network.

\subsubsection{Event-Image Alignment / Synchronization}
\begin{figure}[ht]
  \centering
  \includegraphics[trim={0cm 0cm 0cm 0cm},clip,width=1.0\linewidth]{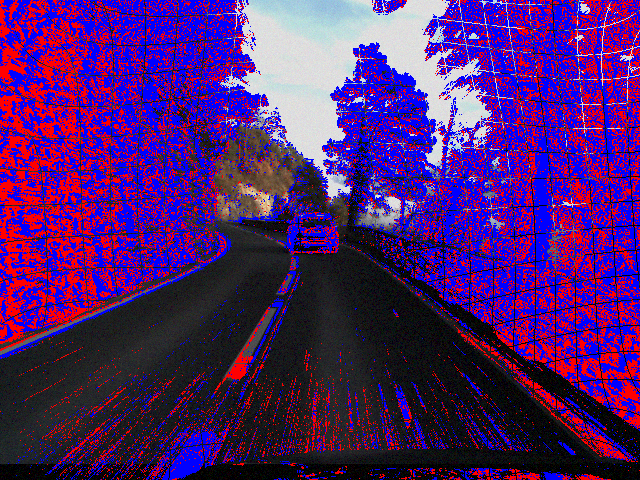}
  \caption{DSEC dataset: real events are spatially aligned to image frames with known calibration parameters. Despite attempted alignment, some misalignment errors and interpolation artifacts are present in the data, affecting HDR reconstruction quality.}
  \label{fig:event_warp}
\end{figure}

Another challenge with our proposed mixed-modality HDR system is the spatial alignment and synchronization of the event and image modalities since they are captured using different camera sensors, with different sizes, resolutions, field-of-views, and with baseline displacement between them. Suppose this system was to be hardware-implemented on a smartphone, for example. In that case, the two sensors could be electronically synchronized, and the baseline between the two optical sensors would be minimal. The network would likely implicitly compensate for any spatial misalignment between the events and images. To show the viability of our proposed system, we presented two cases in our experiments: 1) synthetic events which are perfectly aligned to the image frames, and 2) real events which are misaligned but are spatially warped to the image frames using a calibration checkerboard. In Fig. \ref{fig:event_warp} we show an example of the event-to-image spatial alignment. Note that interpolation artifacts (notice the grid lines across the image) are present in the warped events because of the difference in resolution between the images and events. We find that fine-tuning our models on this dataset helps mitigate these artifacts appearing in the final HDR reconstructions.  

\subsection{Examples from our Datasets}

\begin{figure}[ht]
  \centering
      \includegraphics[trim={0cm 10.4cm 0cm 9cm},clip,width=1.0\linewidth]{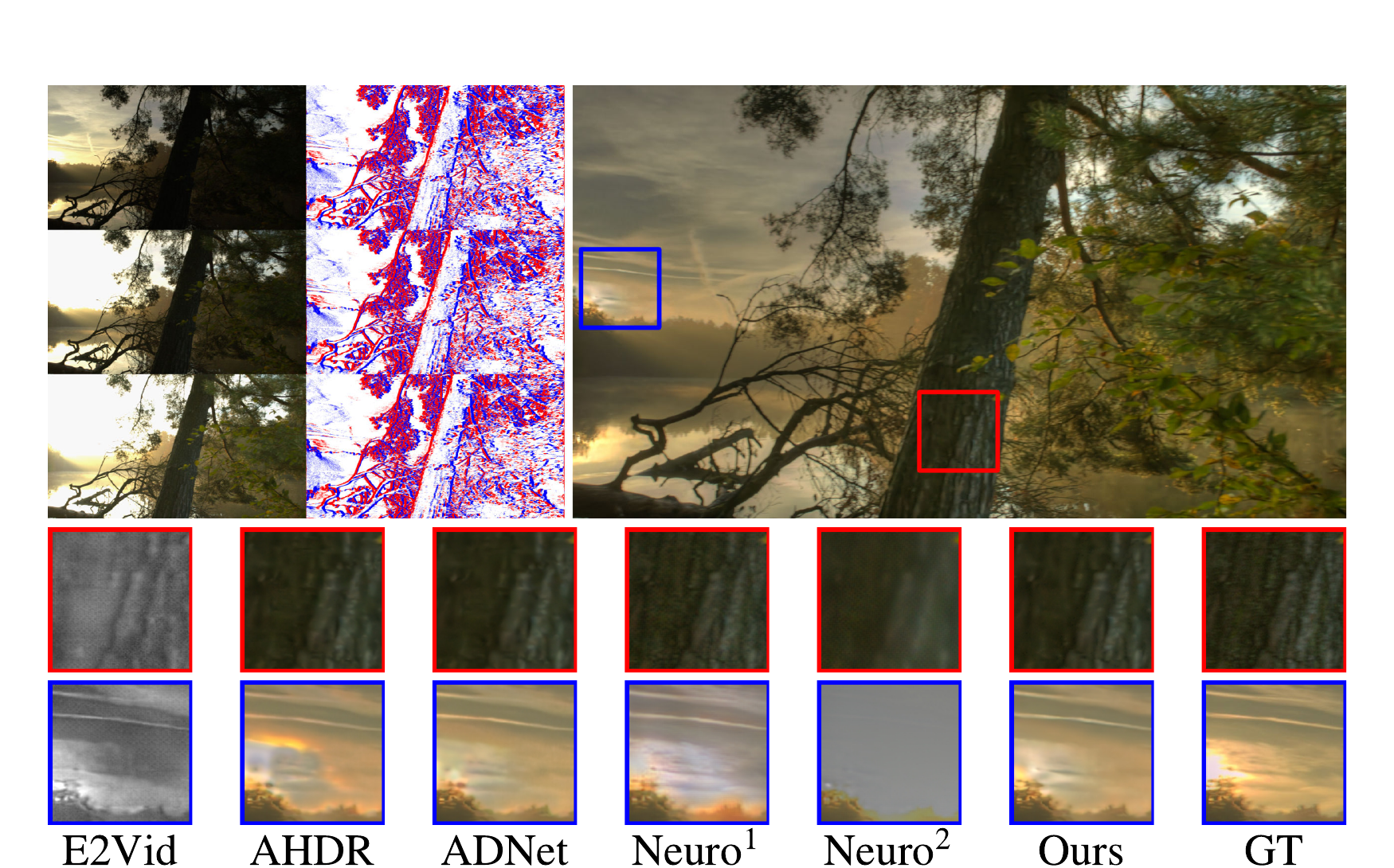}
    \includegraphics[trim={0cm 10.4cm 0cm 9cm},clip,width=1.0\linewidth]{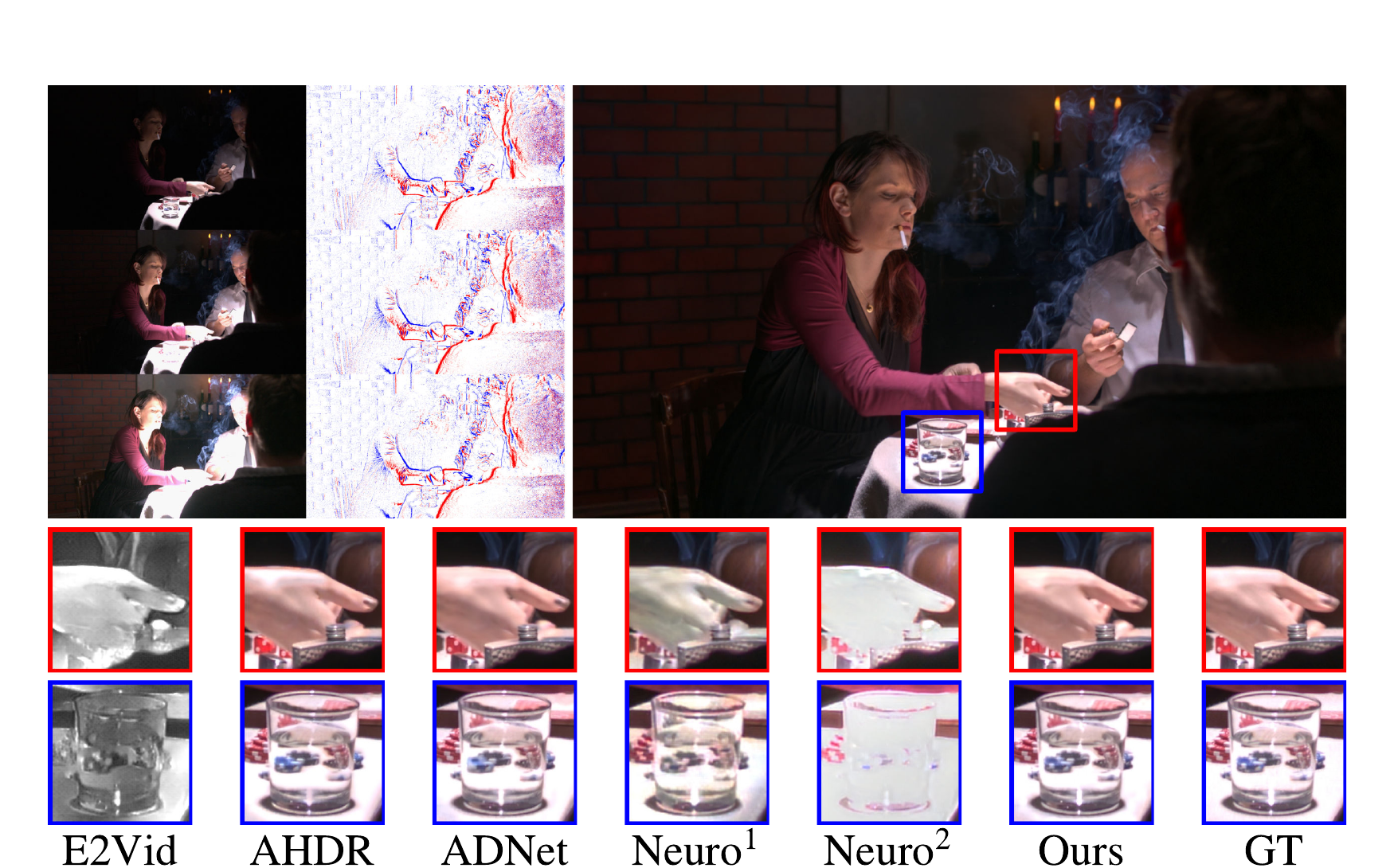}
  \caption{Additional qualitative results on the HdM test set with synthetically generated events. Top left: input LDRs and events, top right: the predicted HDR image using our method, and bottom row: comparison crops for each method.}
  \label{fig:supp_synthetic1}
\end{figure}

\begin{figure}[ht]
  \centering
      \includegraphics[trim={0cm 10.4cm 0cm 9cm},clip,width=1.0\linewidth]{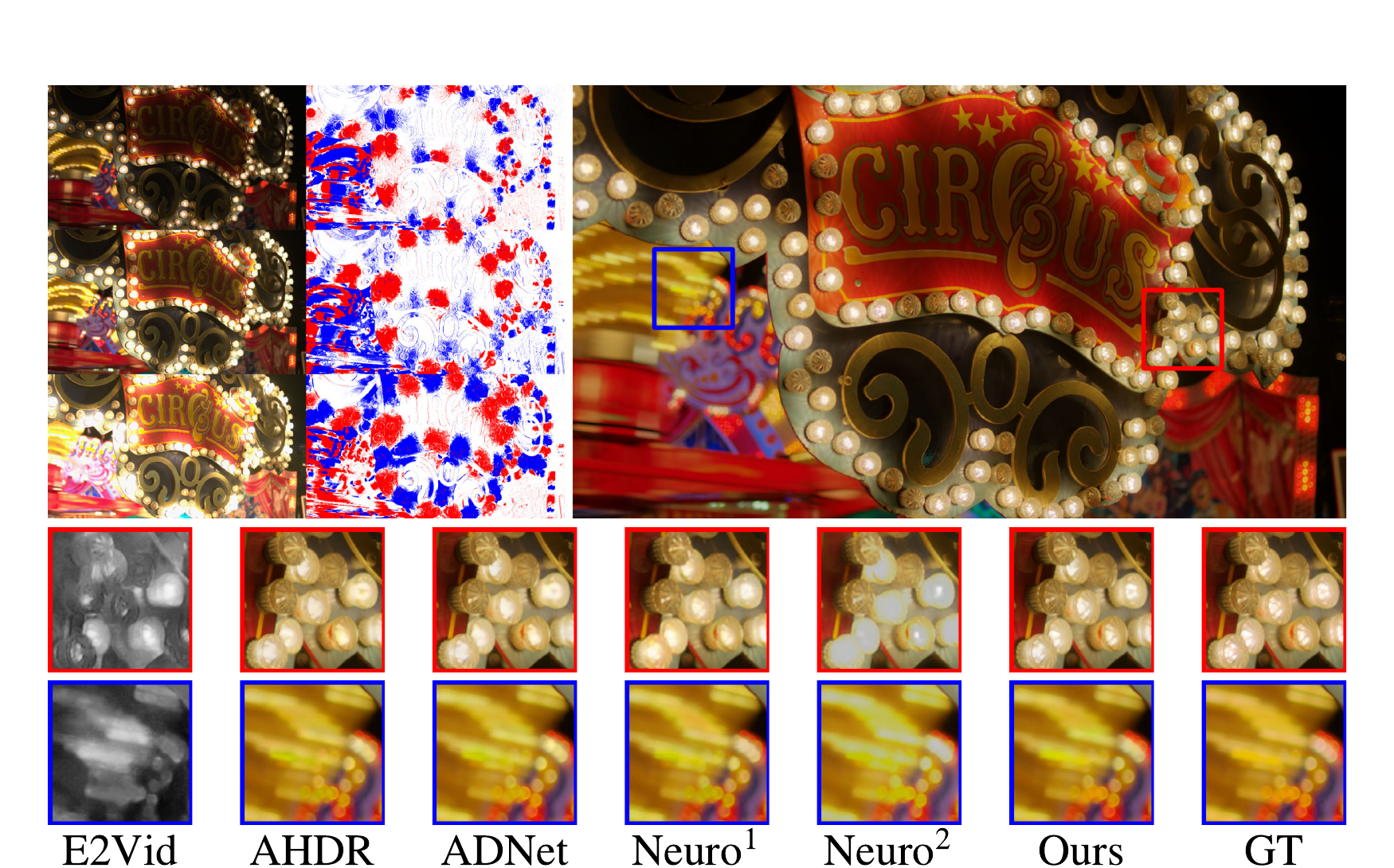}
    \includegraphics[trim={0cm 10.4cm 0cm 9cm},clip,width=1.0\linewidth]{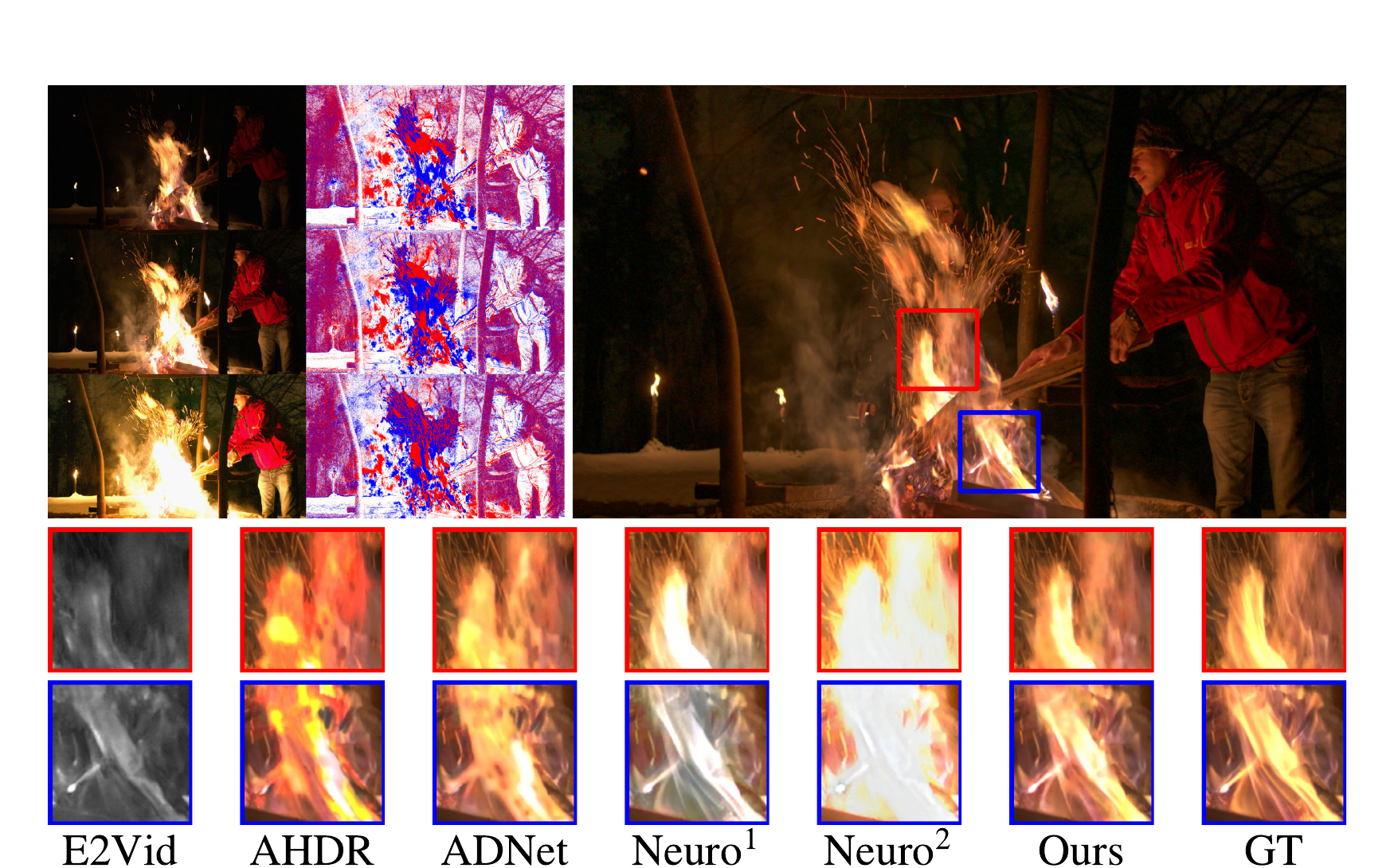}
  \caption{Additional qualitative results on the HdM test set with synthetically generated events. Top left: input LDRs and events, top right: the predicted HDR image using our method, and bottom row: comparison crops for each method.}
  \label{fig:supp_synthetic2}
\end{figure}

\begin{figure}[ht]
  \centering
      \includegraphics[trim={0cm 9.9cm 0cm 8cm},clip,width=1.0\linewidth]{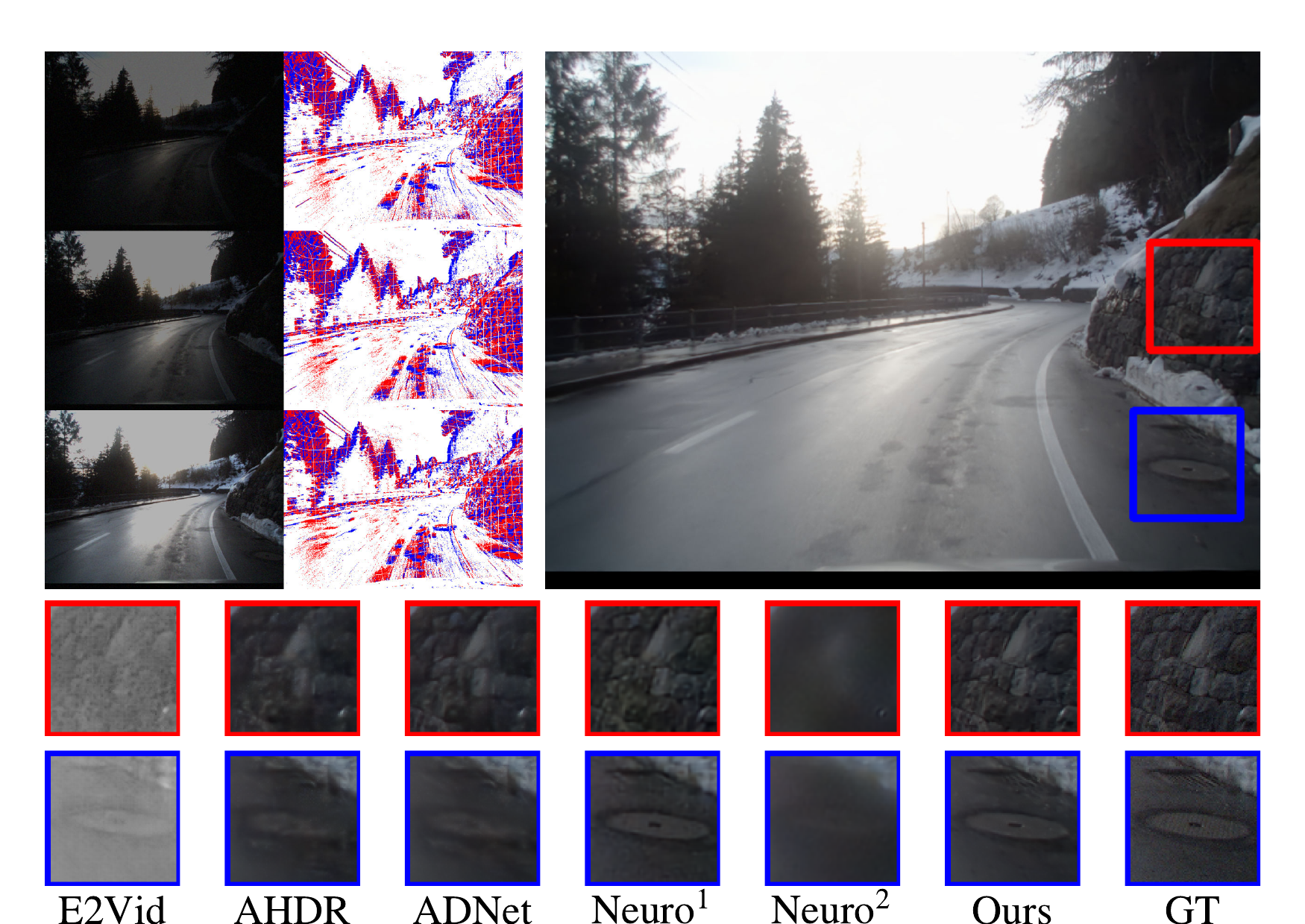}
    \includegraphics[trim={0cm 9.9cm 0cm 8cm},clip,width=1.0\linewidth]{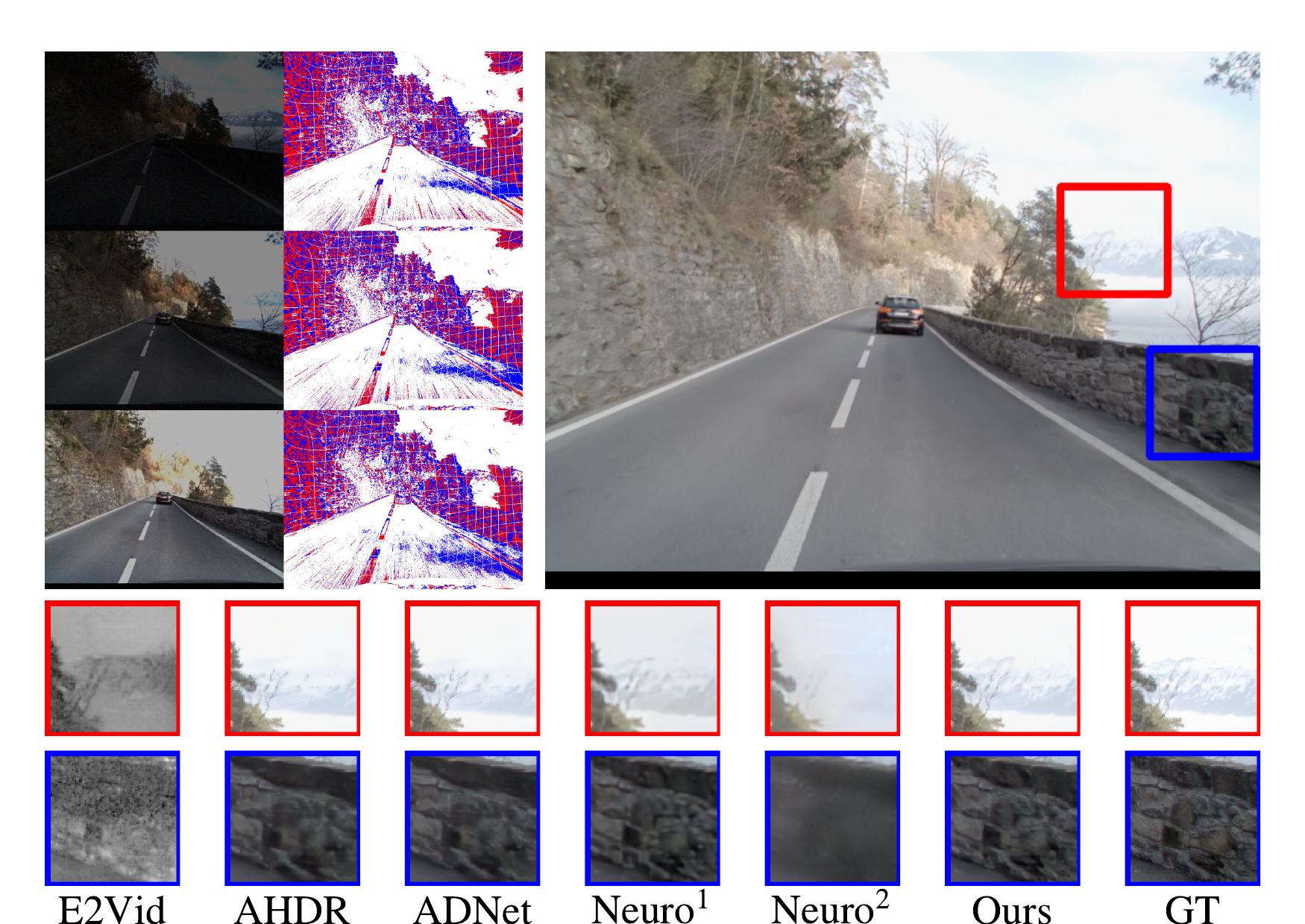}
  \caption{Additional qualitative results on the DSEC test set with real events. Top left: input LDRs and events, top right: the predicted HDR image using our method, and bottom row: comparison crops for each method.}
  \label{fig:supp_real1}
\end{figure}

\begin{figure}[ht]
  \centering
      \includegraphics[trim={0cm 9.9cm 0cm 8cm},clip,width=1.0\linewidth]{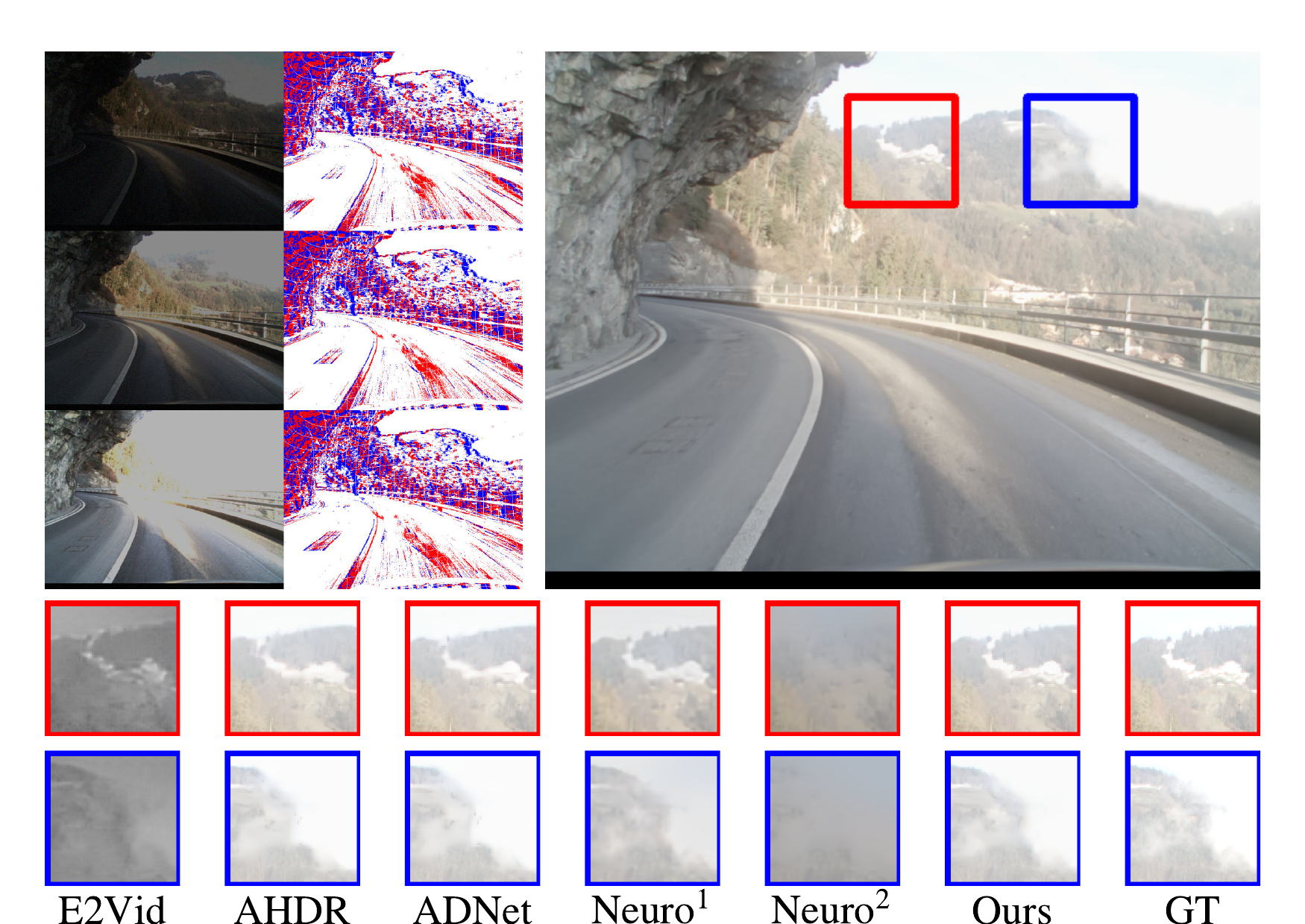}
    \includegraphics[trim={0cm 9.9cm 0cm 8cm},clip,width=1.0\linewidth]{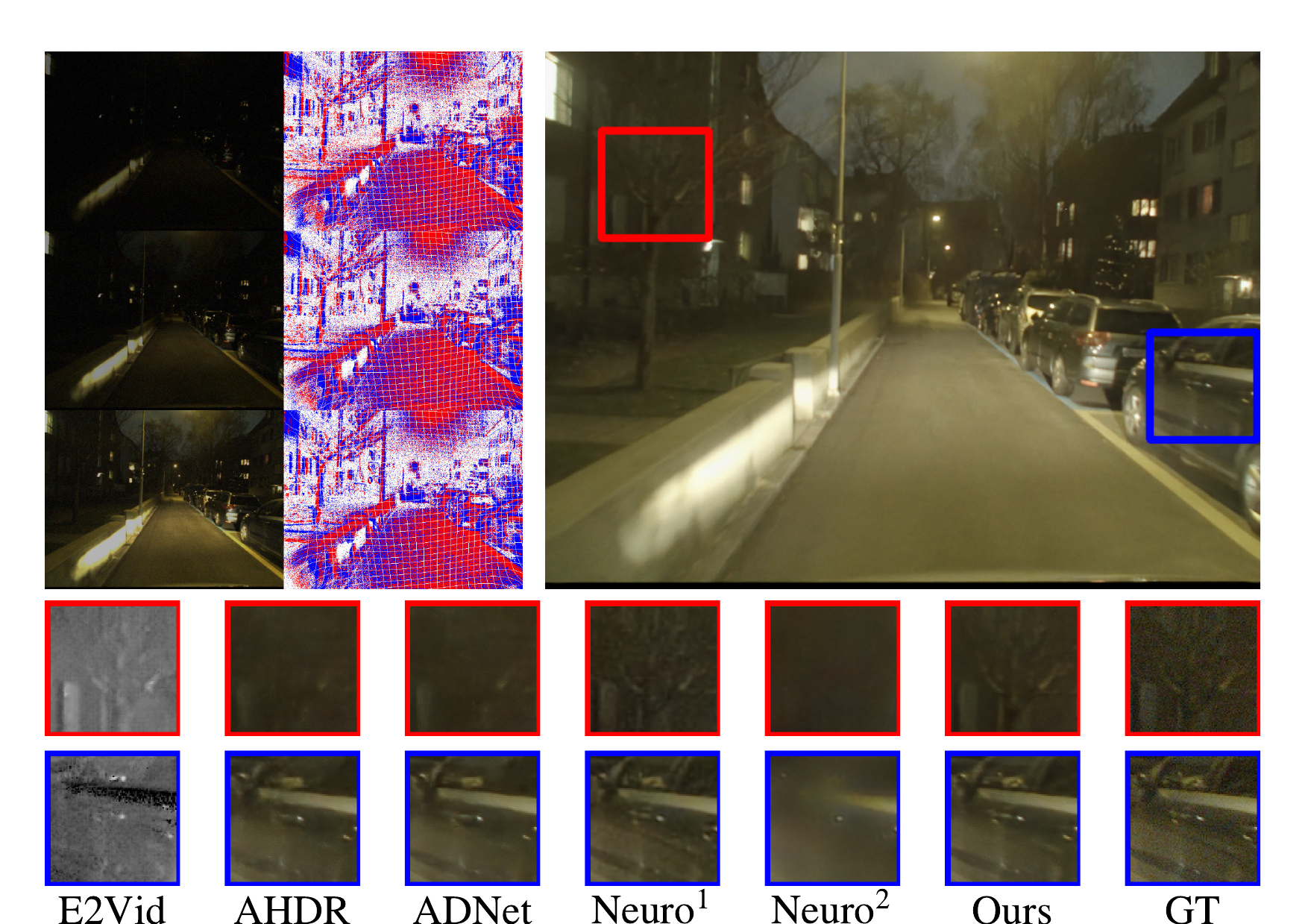}
  \caption{Additional qualitative results on the DSEC test set with real events. Top left: input LDRs and events, top right: the predicted HDR image using our method, and bottom row: comparison crops for each method.}
  \label{fig:supp_real2}
\end{figure}

Here we provide additional examples of our results on each dataset: HDM dataset with synthetically generated events(Fig. \ref{fig:supp_synthetic1} and \ref{fig:supp_synthetic2}), and DSEC dataset with real event data (Fig \ref{fig:supp_real1} and \ref{fig:supp_real2}).


\end{document}